\documentstyle[12pt]{article}
\title{Information Integration\\ and Computational Logic}
\author{Yannis Dimopoulos and Antonis Kakas\\
\small{Department of Computer Science, }
\small{University of Cyprus}\\
\small{75 Kallipoleos Str., P.O. Box 537, }
\small{CY-1678, Nicosia, Cyprus}\\
\small{ \{yannis, antonis\}@cs.ucy.ac.cy}}
\date{}

\begin{document}
\maketitle
\begin{abstract}
Information Integration is a young and exciting field with
enormous research and commercial significance in the new world of
the Information Society. It stands at the crossroad of Databases
and Artificial Intelligence requiring novel techniques that bring
together different methods from these fields. Information from
disparate heterogeneous sources often with no a-priori common
schema needs to be synthesized in a flexible, transparent and
intelligent way in order to respond to the demands of a query thus
enabling a more informed decision by the user or application
program. The field although relatively young has already found
many practical applications particularly for integrating
information over the World Wide Web. This paper gives a brief
introduction of the field highlighting some of the main current
and future research issues and application areas. It attempts to
evaluate the current and potential role of Computational Logic in
this and suggests some of the problems where logic-based
techniques could be used.
\end{abstract}

\section{Introduction}

Over the years a vast and diverse amount of data has been
collected or created and stored by different users for different
purposes. In the last decade it has been realized that new
information can be extracted from this data by synthesizing in an
advanced, sometimes referred to also as intelligent, way the
available information from the initial separate sources. Such
synthesis or {\em integration of information} would facilitate
and enable high-level tasks such as planning and decision making.

The problem of information integration has its origins in the
problems of multidatabases, federated databases and interscheme
extraction for global schema construction. Recently, the need to
address this problem has been further emphasized with the
explosion in the amount of information that is now available
on-line over the information networks such as the Internet and
the World Wide Web. This development has given new dimensions to
the task of information integration requiring the integration of
{\em non-structured}, {\em highly distributed} {\em autonomous}
and {\em dynamic} information sources. A flexible and scalable
strategy for integrating these disparate information sources
while respecting their autonomy is required. At the same time it
has brought with it an enormous commercial potential that is
already helping to drive research in the field.

The general task of Information Integration can be simply stated,
in an informal way, as that of "getting information sources to
talk to each other" in order to support some higher level goal. We
want to achieve integration of information without integrating the
information sources themselves. Thus an information integration
system must provide a  uniform interface to information sources
allowing the user or application program to focus on specifying
what they require rather than specifying how or where to find the
information. Some of the basic technical problems that need to be
addressed by such a system are: modeling of the information
content, query reformulation with flexible selection of
information sources and query optimization. For many of these
problems it is possible to employ, together with database
techniques, other techniques from Artificial Intelligence (AI) and
soon it was realized that the field of Information Integration
lies at the intersection of Databases with Artificial
Intelligence and other areas such as Information Retrieval and
Human Computer Interaction. This in turn has now given an
excellent application domain for (weak) AI.

One of the major difficulties of information integration
stems from the fact that the information sources involved
employ different data models and are heterogeneous both
semantically and syntactically. In addition, it is envisaged
that information integration systems would be able to support in an
integral way advanced applications exploiting the richness of the
information made available by these systems.
These two factors necessitate rich formalisms for semantic interpretation
of the available information so that it can be understood by the other
sources and the application programs. We are thus led to the need for
{\em semantic (or intelligent) information integration} where, along
with techniques from databases, Artificial
Intelligence and Computational Logic can play a significant role
in providing explicit and declarative representation frameworks for
modeling the information and the complex behaviour of the system.
The autonomous nature of the information sources, where we may only be
able to represent partial information and where information from
different sources may be overlapping or even inconsistent, increases
the possibilities to apply techniques from the
areas of Artificial Intelligence and Computational Logic.

In particular, a declarative logical representation can form the
conceptual basis for the basic architecture of Information
Integration systems. The primary role of Computational Logic
transpires to be that of providing an overall cooperation layer
that would link, through suitable forms of inference, the
different information sources into an intermediate layer. This
intermediate layer is called in the Information Integration
literature, a {\em mediator}. Within this logical framework it
would also be easy and natural to employ specific Computational
Logic methods for particular subtasks required inside an
information integration system.
%
A {\em hybrid} computational model emerges where logic and its
reasoning offer the computational link between high-level complex
requirements and lower-level (typically non-logical)
computational tasks such as searching, constraint solving,
retrieval and various forms of manipulation of data.

The rest of the paper is organized as follows. Section 2, briefly
reviews the main elements of Information Integration Systems and
the current role of logic in them. Section 3 presents some of the
Information Integration systems which use (to a varying extend)
logic and discusses the current scope of application of such
systems. Section 4 presents some of the future problems in the
area and the challenges for Computational Logic stemming from
these. The paper concludes with a short appraisal of the
potential role of Computational Logic in the area.

\vspace*{.4cm} \noindent {\bf Disclaimer:} This paper is a survey
written mainly for the non-specialist in the field of Information
Integration and does not claim to provide a complete account of
the area and its particular systems and applications.
Furthermore, the presentation of the paper follows a particular
view concerned with the possible role of Computational Logic in
this new field attempting to show links between existing work in
Computational Logic and problems arising in the area of
Information Integration. In some cases, these potential links
reflect the authors personal opinion.

\section{State of the Art}

The first roots of Information Integration can be found in the
area of databases with the problem of constructing {\em
mutlidatabases} \cite{multiDB1,multiDB2,multiDB3}. This aspired
to form databases where different sites of heterogeneous databases
would cooperate and be updated together. In order to reduce the
complexity of the problem the notion of {\em Federated Databases}
\cite{fedDB1,fedDB2,lro} was proposed and later on {\em Data
warehousing} \cite{dataware1,dataware2,dataware3} helped to
simplify the problem even more by storing data in one site.

\subsection{The Basic Architecture}

To address the advanced requirements of future information
systems, Wiederhold proposed \cite{wie92} a general system
architecture where a central role is assigned to software modules
called {\em mediators}. A mediator is an information system
component that lies between the data sources and the user and her
applications. A mediator "exploits encoded knowledge about the
data to create information for a higher layer of application".
Its added value is exactly this transformation of data to
information. The tasks of a mediator according to \cite{wie97}
include:

\begin{itemize}
\item accessing and retrieving relevant data from multiple heterogeneous
resources,
\item abstracting and transforming retrieved data into a common representation
and semantics,
\item integrating the homogenized data according to matching keys, and
\item reducing the integrated data by abstraction to increase the information
density in the result to be transmitted.
\end{itemize}

The mediator-based architecture has been to a large extend
adopted in the development of the reference architecture for
Intelligent Information Integration (I3) supported by DARPA
\cite{darpa}. More importantly, the need for an intelligent
middleware between the sources and the user is a common
assumption in most of the work in I3. As already noticed in
\cite{wie92}, to perform their functions effectively, mediators
will embody techniques from logical databases and AI. It is
inside the mediator, in its design and implementation, where
logic can play an important role in the information integration
systems of the future.

\subsection{Inside the Mediator}

Among the above mentioned tasks of the intelligent mediator
middle-ware, the first, often called the {\em information
gathering problem}, is the most studied. This task can be
decomposed in the following sub-tasks.

\begin{itemize}
\item modeling the contents of information sources
\item modeling the relation between sources and the mediator
\item query reformulation (query planning)
\item query optimization and execution
\end{itemize}

\subsubsection{Modeling and Query Reformulation}

As mentioned above a mediator needs to support adequately a wide
range of tasks and provide a variety of functionality based on
these tasks \cite{darpa,wie97}. In order to achieve this
mediators need to use models of the various information sources
that are available, referred to as the {\em source model}, as well
as models of the user큦 (or the application큦) view of the problem
world, referred to as the {\em world model}, and to link these
models together.
%
Among these modeling tasks of a mediator the linking of the world
model to the source model is of central importance.
In most of the early studies for developing such models the
simplifying assumption is made that the sources can be modeled as
relational databases. Similarly, it is assumed that the world
model is captured by a set of relations. These relations are not
stored in any of the information sources, but are those on which
the user (or application program) will pose queries. In DB
terminology, we say that the world's model relations define a {\em
global mediated schema}.

Roughly speaking, the relations in the mediated schema are used
by the user (or her/his application program) in order to specify
{\em what} is the information s/he wants, while the source
relations determine {\em where} the information is stored. To be
able to answer user queries the mediator needs to know how the
global schema relations are related to those of the information
sources. Having this knowledge the middle-ware of an information
integration system translates queries on the mediated schema to
queries on the source relations.


There are basically two approaches to relate sources to the mediated schema.
The first is the {\em local as view}  approach and the second the
{\em global as view} approach \cite{flm} (also called {\em source definition} and
{\em view definition}  respectively ).
In the first approach sources are considered to be materialized
views over the mediated schema. Datalog programs of the form
\begin{verbatim}
source_relation :- global_relation_1, ..., global_relation_n
\end{verbatim}
are used to link source relations to the mediated schema
describing in this way the contents of the source in terms of the
global mediated schema. A datalog rule like this is called a {\em
conjunctive query} expressing the fact that the query given by the
body of this rule can be answered using the source relation found
at the head of the rule. More generally, query translation can be
seen as synthesizing queries from {\em views} \cite{ullman97}, a
problem that has been studied quite extensively
\cite{ad,dg,dl,lk,lmss,lro}. Levy \cite{levyreview00} provides a
thorough survey of this problem of answering queries using views
in different application areas, including information integration.

In the global as view approach the mediated relations are considered
as views over the source relations. Rules of the form
\begin{verbatim}
 global_relation :- source_relation_1, ..., source_relation_n
\end{verbatim}
relate the source relations to the global relations. Query
translation then takes the form of simple view unfolding.


Ullman in \cite{ullman97} compares the two approaches, and
identifies their advantages. The global as view approach makes
the query translation problem easier. It can also draw from work
on abduction \cite{KakasSurvey}, when we want to increase the
expressive power of the modeling language to include constraints,
negation and other powerful modeling features. On the other hand,
the local as  view approach simplifies adding and deleting
sources and therefore it is particularly useful in dynamic
environments. Moreover, it facilitates the specification of
constraints on the content of the sources as we can attach such
constraints in the conjunctive query that describes them. In the
global as view approach constraints on the contents of a source
can be expressed separately as meta-level integrity constraints
on the source relations.

Among the existing system TSIMMIS \cite{tsimmis1}, Disco
\cite{disco}, and Coin \cite{coin1,coin2,coin3} adopt the second
approach while Information Manifold \cite{lsk,lro} and Infomaster
\cite{info1,info2} follow the first. Moreover, recently there
have been systems that combine the two approaches
\cite{flm1,picsel}. We note here that description logics are also
used as the underlying formalism for mediator modeling in some
information integration systems, including SIMS \cite{sims} and
PICSEL \cite{picsel}. The general role of description logic in
information integration has been studied in
\cite{catarci,lenzerini3,lenzerini5}, while work on answering
queries using views in the context of description logic is
described in \cite{beeri,lenzerini4,rousset-dl00}.

Once the mediator translates user queries to sources relations,
it has an {\em information gathering plan}, which in abstract
terms can be seen  as a conjunction over the source relations. At
this stage the mediator knows what are the information sources
that it needs to access in order to answer the query. Then it must
retrieve the information form the sources by sending requests to
them in a language that they understand. More importantly, the
mediator should not post to the sources queries that are beyond
their answering capabilities.  Consequently the mediator must also
model the {\em query constraints} of the sources. An important
family of such constraints are {\em binding restrictions}, i.e.
information that specifies that a certain source can only answer
queries that have some of their attributes bound. Such query
processing systems that take into account source capabilities are
described in \cite{flms,gly,pgh,rsu,vp,vp1,ylgu,yl}.

Another important feature of a mediator  at this stage is its
capability to model various forms of {\em local completeness} of
the sources, that is expressing the extend to which a source is
complete for the domain that it covers. This information can help
the system to restrict and improve access to the sources
\cite{etzioni,razor,duschka97,lk,levy96}.

\subsubsection{Query Optimization and Execution}

Apart from being {\em sound} (the answers returned to the user
are correct) and {\em complete} (all correct answers to the query
are found) information gathering plans need also to be efficient.
Query optimization in the context of information integration has
several aspects. These can be divided into {\em logical
optimization} and {\em execution optimization}. Logical
optimization \cite{lk}, seeks to eliminate redundancies from the
information gathering plans possibly by exploiting local
completeness information \cite{etzioni,razor,duschka97,lk,levy96}.
Like in a traditional DB system, the resulting logically
optimized plan is passed to a query optimizer that has to come up
with an efficient query execution plan to achieve execution
optimization. While this is a well-studied problem in the relation
DB theory, information integration adds extra complexity
\cite{lk,levyrev99}.

Garlic is one of the early integration systems that address
logical and execution optimization by extending previous work in
relational DB \cite{garlic1}. More recent work in the field is
concerned with using local completeness and source overlap
information  for ordering source accesses in order to maximize
the likelihood of obtaining answers as early as possible, and to
minimize the access cost and the network traffic
\cite{levyprob,lk,vp2}. Another way to consider optimization is
to add new sub-queries in the query plan specifically for the
purpose of optimization \cite{ecp97}.

The dynamic nature of the environment in which information
integration systems operate require that their query execution
engines have advanced capabilities. Information integration over a
network like the Internet has to take into account that sources
can become slow or unavailable during the execution of a
information gathering plan. Query planning needs to be interleaved
with query execution, and the system must be able to re-plan and
re-optimize \cite{knoblock,ifflw}.
%

\subsection{Source Structure and Heterogeneity}

The previous discussion makes the simplifying assumption that the
contents of the information sources can be described by a set of
relations,
However, there are two separate problems with this approach
\cite{lreview}. The first is that sources can be {\em physically
semistructured}, where the structure of the source information is
obscured by the way that this information is stored e.g. in HTML
file sources often the data is not stored directly in its
structured form due to additional mark-up information used to
make the source human readable.
On the other hand data can be {\em logically
semistructured} in the sense that the data does not necessarily
fit in a schema. Data can be so irregular that the schema can be
very large.

To cope with physically semistructured sources we add {\em
wrappers} to them, while the problem of integrating logically
semictructured sources is tackled through flexible data models
and query languages. We discuss these important issues in the
following two subsections.

\subsubsection{Wrapper Technology}

Sources often do not provide their data in a format, eg.
relational tuples, that can be directed manipulated by an
integration system. For example, if the source is an HTML page, it
can be the case that structured data (eg. names, phones and
emails of the faculty of a university department) is embedded in
natural language text and graphical presentation objects. In such
cases the integration system communicates with a {\em wrapper},
i.e. a program whose task is to translate the data in a form that
can be further processed by the integration system, eg. by
removing all HTML markup information. Since sources usually store
their data in different formats, each source often needs a
different wrapper. However, writing wrappers can be a tedious
task, therefore {\em wrapper generation} is an important issue.
Wrapper generation can be {\em semi-automatic}, when support
tools are used in their generation, or {\em automatic}, using
machine learning techniques. The work of
\cite{eikvil,muslea-intro} surveys the field and discuss the
relation of wrapper generation to the more general problem of {\em
information extraction}\footnote{An information repository on
information extraction and wrapper generation is maintained at
http://www.isi.edu/~muslea/RISE/.}. We should note however that
the emergence of XML as a new standard for representing data in a
machine readable way (in contrast to HTML which focuses on human
readable representations) is expected to significantly mitigate
the problem of physical data heterogeneity.

\subsubsection{Semistructured Data}

There are many interesting cases where the data can not be
constrained by a rigid schema. Moreover, in an information
integration application where data needs to be exchanged and
transformed among sources with different data models (eg.
relational and object oriented) we require very flexible data
formats. These needs have led to the new research problem of
modeling and querying {\em semistructured data}
\cite{buneman,abiteboul,suciureview}. While there is no strict
definition of the term, semistructured data is often described
as  {\em schemaless} or {\em self-describing'} data, ``terms that
indicate that there is no seperate description of the type or
structure of data'' \cite{abibook}. Semistructured data is
captured by flexible data models like the OEM model of the TSIMMIS
project \cite{tsimmis1} or, more abstractly, by some form of {\em
edge labeled directed graphs} \cite{buneman}. All schema
information moves to the data itself, i.e. to the graph, in the
form of the labels attached to the nodes or edges of the graph
(hence the term self-describing).

Flexible query languages like Lorel \cite{lorel1,lorel2} and UnQL
\cite{unql} provide several required functionalities such as (i)
the ability to query without knowing the exact structure of the
data,  (ii) the provision of navigation queries in the style of
Web browsing and (iii) the ability to query both the data and the
schema at the same time. The problem of answering queries using
views in the context of semistructured data has also attracted
attention recently \cite{CGLV99,CGLV00-1,CGLV00-2,yannis-semi-re}.
Motivated by the importance of schema information for efficient
processing, storage and usability of data, recent research
attempts to bring back some forms of structure or schema to
semistructured data. Several schema formalisms have been
considered in different approaches to this problem of extracting
schema from semistructured data, eg. in
\cite{bunemanstr,dataguide,nestorov,nestorov1}. Logical
approaches to schema formulation from semistructured data include
the use of Datalog rules \cite{NAM}, description logics
\cite{lenzerini1} and mappings into the relational database model
\cite{STORED}.

The semistructured data model represents a new paradigm and
challenge in databases and therefore in Information Integration.
The database community has already made significant progress
towards linking semistructured data research to emerging
standards for information exchange like XML
\cite{XML,XML-challenge}. For instance, several of the query
languages developed for XML \cite{XML-QL,XQL,XML-GL,XQL-compare},
bare a strong resemblance to query languages for semistructured
data. A recent book \cite{abibook} offers an extensive coverage
of this research theme.

\subsection{The Semantic Web}

Recently, it has been realized that information integration over
the Web can benefit significantly by adding extra semantic
structure over these information sources, namely the web
documents or web pages. This need for semantic information links
information integration with the {\em Semantic Web}, a term coined
by Tim Berners-Lee \cite{SW-roadmap}. The Semantic Web is a
vision for the next stage of the Web, described as a space of
self-describing, machine-processable or "machine-understandable"
fragments of information in which documents convey meaning. In
contrast to the existing Web, whose content information is only
human-readable, the Semantic Web aims to develop the technologies
that will enable the machines themselves to make more sense out of
it.

The Semantic Web\footnote{See www.semanticweb.org for a variety of
information about the Semantic Web.} is envisaged to have a
multi-layer architecture which is based on the XML standard for
semi-structured Web-Data and related technologies (eg.
Namespaces, XML-Schema) that support inter-operability mainly on a
syntactic level. On top of XML lies the Resource Description
Framework (RDF) \cite{RDF},  a key concept of the Semantic Web.
Essentially, RDF is a simple data model for expressing {\em
metadata} in the form of object-property-value triples, called
statements. In RDF statements can be the object or value of a
triple, allowing in this way the nesting of statements.

While RDF provides the model for describing resources and
relationships between them, it does not support any mechanism for
declaring properties or relationships between properties and other
resources. For this {\em RDF Schema} (RDFS) \cite{RDFS}, a schema
specification language, is used. RDFS provides a basic type
system, or vocabulary, for use in RDF models. Important modeling
primitives it provides include class/subclass, instance-of, and
range/domain restrictions. However, we should note that W3C
recommendations follow a minimalistic approach, so neither RDF
nor RDFS provide any formal semantics for the constructs they
define, nor do they provide an {\em inference system} or a query
language. This lack has motivated a number of works that attempt
to relate RDF and RDFS to formalisms with clear semantics like
conceptual graphs \cite{CGRDF} or logical languages
\cite{que-inf-rdf,boley,web-formal,log-inter,metalog}.

Hence although RDF/RDFS provide a powerful infrastructure for
semantic inter-operability, they do not suffice to capture the
advanced knowledge representation and reasoning needs of the
Semantic Web. The Semantic Web architecture requires two
additional layers, an ontology and a logic layer, that aim at
accommodating richer forms of knowledge. One of the early
attempts to employ ontologies on the Web were Ontobroker
\cite{ontobroker} (that will be described in some detail in the
following)  and the SHOE project \cite{shoe} that allows authors
to annotate their HTML pages with ontology-based knowledge about
the page contents. More recent approaches are  the {\em Ontology
Inference Layer - OIL} \cite{OIL} and the {\em DARPA Agent Markup
Language - DAML} \cite{DAML}. OIL \cite{OIL1,OIL2} is a system
that combines the reasoning capabilities of description logics
(more specifically that of FaCT \cite{fact1,fact2}), with the rich
modeling of frame-based systems and the new Web standards of RDF
and XML. The DAML Program \cite{DAML1,DAML2}, which officially
began in August 2000, is an ``... effort to develop a language and
tools to facilitate the concept of the semantic web.''. The DAML
language also builts on top of XML and RDF, and its latest
release called DAML+OIL is available from \cite{DAML+OIL}. Two
recent papers \cite{IEEE1,IEEE2} provide an up-to-date overview
of languages, tools and services of the Semantic Web, and discuss
new developments and current issues.

\subsection{The Current Role of Logic}

One of the basic requirements of Information Integration
frameworks is to allow the user to specify {\em what} information
she wants rather than {\em how} to get it. Such systems must
rely, at least at some level of abstraction, on a declarative
representation with a rich formalism for describing the available
information. Logic can provide such a flexible representation and
thus play an important role in the development of information
integration systems.
%
In the emerging future systems where an advanced form of sythesis
of information would be facilitated through an application layer,
above the mediator layer, Logic and Artificial Intelligence will
have an ever increasing role to play. This wider role of AI in
current and emerging Information integration systems has been
recently presented in the survey paper \cite{levyrev99}.

Currently, the central role of logic rests on the fact that it
provides a framework for specifying the mediator layer of an
Information Integration system and the relation between the
mediators and source information. Logic as a Knowledge
Representation formalism facilitates this and provides (in some
cases) techniques for the computational process of the Information
Integration system. Most systems indeed either use logic
explicitly (e.g. COIN or Infomaster) as a KR technique or can be
cast in to this e.g. the MSL language of the TSIMMIS project. The
work of \cite{ullman97} provides a survey of this role of logic
showing how both the global as view  and the local as view
approaches for modeling the information sources and their
relation to the mediator layer can be understood in terms of
Logical views over Datalog programs. The extraction and
integration of information by these systems is then seen as
answering conjunctive queries over these programs.

A third approach to modeling the information in an Information
Integration system is through the use of Description
Logics\footnote{Description logic has also been used
\cite{calabria-scheme-3} in the related problem of interscheme
extraction from database schemes.}, (e.g. in Information Manifold,
PICSEL, SIMS). This is used as a richer representational
framework that can capture additional (meta-level) knowledge on
the sources available primarily as contraints on the information
in these sources. For example, the {\em PICSEL} system
\cite{picsel} which is based on the Description Logic language
{\em CARIN} is able to combine the global as view  and local as
view  approaches and to use the generic inference mechanisms of
this logic. In particular, it avoids the problem of reformulation
of the query required in the local as view  approach.

At the level of computational methods used in information
integration systems the problem of query answering can be
directly related to logical inference for the reduction of the
query to the mediator layer (often this is trivial) and then the
further reduction of this to the information sources. Indeed, in
the global as view  approach one way to formalize this reduction
is as a simple form of abductive reasoning \cite{tonyvldb}. In
the local as view approach this can be seen as a problem of
reformulating a query in terms of (materialized) views and in turn
this is closely related to the problem of query containment in
Datalog. Logic helps to study the theoretical issues concerning
the question to what extend it is possible to do this
reformulation. In Infomaster this query reformulation is achieved
via first inverting the local as view description of sources in
terms of mediator relations and then reducing, as in the global
as view approach, the original query using logical inference
techniques to do this reduction. Note that this inversion can
bring us outside datalog with disjunction and recursion required.

Deductive databases with their strong logical basis can also play
an important role in information integration, at the analysis and
design level and also in the actual implementation. Indeed,
Datalog-based formalisms are used by most approaches that build
integrated systems and mediators for information taken from
multiple heterogeneous data such as legacy databases, external
views, and web sites \cite{levyrev99, levyullman99}. Furthermore,
the  query optimization problem for these systems is frequently
solved by means of annotated plans where rules and goals are
assigned bound/free annotations. This is the very approach
developed and used by deductive database compilers and optimizers
\cite{rsu, levypatterns99}. The Deductive Database system of LDL++
\cite{ldl++} has been used as a component in the (early)
information integration systems of Carnot
\cite{carnot,ddbs-carnot95} and Infosleuth \cite{sleuth,
sleuth2}. LDL++ also  proved very useful in an assortment of
other tasks needed for information integration, such as the task
of cleaning and converting legacy data \cite{dbs-use95}.

In approaches that use description logics query processing is
based in a general way on the underlying inference algorithm of
the logic. This has the advantage that the development of a new
application domain amounts to the design of a suitable knowledge
base without the need to develop a specialized query engine. The
description logic approaches also have a high degree of
modularity. This stems from the fact that the representation of
the sources is separated into two parts: (i) a theory that
describes the relation between mediator and sources and (ii)
declaratively stated integrity constraints that capture additional
knowledge about the information in the sources.

The COIN framework \cite{coin1} adopts a similar approach for its
representation of the domain using integrity constraints to help
resolve semantic conflicts between information that can be
acquired from different sources, thus giving a form of semantic
query optimization. This system is implemented on the ECLiPSe
parallel constraint logic programming environment and relies on
techniques of abduction and constraint propagation.

It is important though to note that despite the strong logical flavour
that some of these information integration frameworks have many of the
techniques that are used in these systems are specific to the
particular needs of the systems. In particular, the problems of
query optimization, source capabilities, local completeness of sources
and use of other meta-level information about the source information
are typically handled with specific techniques that vary form system to
system.

Some of these techniques e.g. for query planning, take input from
AI and Logic (and Deductive databases) but there are no general
methods of how logical techniques can be exploited in these
important aspects of the general problem of information
integration. Computational Logic needs to develop its own
specific techniques suited for the particular problems of
information integration. These techniques need not be (and in
many cases should not be) entirely logical but they should have a
hybrid nature of co-operation between logical reasoning and other
computational methods.

The working assumption of much of the early research is that the
information stored in the sources can be captured by a set of
relations, i.e. it is assumed that sources are relational and can
be described by some rigid schema . This assumption is not valid
when working with semistructured data \cite{abibook,lorel1,
calabria-ssd-2} and there is a current strong shift away from
this assumption. It is therefore important for logical approaches
to be able to adequately represent semistructured information
sources. There are some studies towards this direction, where a
logical theory is used to represent the labeled graph data model
for semistructured data. The FLORID approach \cite{bertram},
described in more detail below, uses F-logic, a logical framework
that combines techniques from object-oriented databases with the
power of deductive rules for expressing complex queries. In
\cite{lenzerini1,lenzerini2} the authors propose description
logics as a logical formalism capable of capturing the labeled
directed graphs model of \cite{buneman} and extending it with
various forms of constraints and the capability to deal with
incomplete information. Finally, the MOMIS \cite{momis} approach,
that also builds on work in description logics, also provides
features that allow the representation of some forms of
semistructured data.

\section{Information Integration in Practice}


Information integration has important practical applications
particularly in view of the emerging need to integrate information
distributed over the WWW. Several practical systems of Intelligent
Information Integration have been developed. General information
on these can be found at the following URL addresses
\cite{TZI-DE,DUSKA-Projects,UNI-KARL-DE}. We present here some of
these  systems whose development has to a certain extend been
influenced by Computational Logic either at the conceptual and/or
the implementation level. Other systems that make use of methods
from Computational Logic but to a lesser degree include the early
systems of Infomation Manifold \cite{lreview}, TSIMMIS
\cite{tsimmis1}, SHOE \cite{shoe}, SIMS \cite{sims} and WHIRL
\cite{whirl}.

\begin{description}

\item [COIN] (COntext INterchange) \cite{coin2,coin3}
is an Information Integration framework with emphasis on
resolving problems arising from semantic heterogeneity, i.e.
inconsistencies arising from differences in the representation
and interpretation of data.  This is accomplished using three
elements: a shared vocabulary for the underlying application
domain (in the form of a  domain model), a data model (COIN), and
an application/query language (COINL). Semantic inter-operability
is accomplished by making use of declarative definitions
corresponding to source and receiver contexts i.e. constraints,
choices, and preferences for representing and interpreting data.
The identification and resolution of potential semantic conflicts
involving multiple sources is performed automatically by the
context mediator. Users and application developers can express
queries in their own terms and rely on the context mediator to
rewrite the query in a disambiguited form.

The COIN data model is a deductive and object oriented model of
the family of F-Logic \cite{F-logic}. The mediation engine's main
inference is implemented by means of a resolution based abductive
procedure in the Constraint Logic Programming language of ECLiPSe
\cite{eclipse}. Queries and subqueries are represented by the
successive states of a constraint store along one branch of the
resolution tree. Integrity constraints are translated into
Constraint Handling Rules of ECLiPSe. Their propagation achieves
a form of Semantic Query Optimization by rewriting the queries
and subqueries in the store in between the resolution steps and
by pruning the rewriting process. COIN has been used in various
applications of e-commerce.

\item [FLORID] (F-LOgic Reasoning In Databases)
 \cite{bertram} is a logic-based system for
information extraction and integration from the Web.  Web
exploration, wrapping, mediation, and querying is done in a
monolithic system where F-Logic serves as a data model and a
uniform declarative programming language for Web access, data
extraction, integration, and querying. The Web and its contents
are regarded as a unit, represented in an F-Logic data model.
Based on the Web structure, given by its hyperlinks, and the
parse-trees of Web pages, an application-level model is generated
by F-Logic rules.  For this, the F-Logic language is extended
with Web access capabilities and structured document analysis.  By
retaining the declarative semantics of F-Logic also for the Web
interface methods, Web data extraction can be programmed in a
clear and natural way.  In particular, generic rule patterns are
presented for typical extraction, integration, and restructuring
tasks \cite{WWWCM-99}, such as HTML lists and tables and
syntactical markup.

Web access, wrapper and mediator functionality, restructuring,
and querying can be arbitrarily combined, and thus FLORID can be used
both for Web querying and for information extraction.
The combination of Web access, Web data extraction, and
interpretation rules allows for data-driven Web exploration: a
priori unknown Web pages can be accessed and evaluated dependent
on previously extracted information.  Equipped with suitably
intelligent evaluation rules, the system can explore hitherto
unknown parts of the Web, coping with the steady growth of the
Web.  The practicability of FLORID has been shown in the case
study \cite{MONDIAL} where geographic information from several
sources on the Web has been integrated.

\item [HERMES]
operates on the philosophy that the role of logic in
applications is more that of providing a facilitator for cooperation
between computational models rather than performing the
actual computations itself. Logic is used to integrate computations
in classical datastructures with specialized data
structures for computations been key to scalability.

In HERMES, external databases and software packages are assumed to
either have a legacy Application Program Interface (API) (or have
one built for them), and these sources are accessible via their
API functions. HERMES mediators contain syntax to execute
operations in these packages, and to return the answer in a set.
HERMES rules are expressed in logic, and they allow us to execute
boolean combinations of these API function calls.  HERMES' ``base
predicates'' are just such boolean combinations of API calls --
``derived predicates'' may be defined (recursively or otherwise)
in terms of these ``base'' predicates.
An HERMES query may involve accessing data in a distributed
environment from a RDBS, a logistics data base, a GIS, a route
planning system and a linear optimizer. Over the years, HERMES
has been used to integrate  Oracle, Ingres, and Quad-tree
databases, route planners, logistics databases, Dow Jones stock
mediators, and intelligent travel agents.

\item [Infomaster]
An essential feature of Infomaster \cite{info1} as an Information
Integration framework is its emphasis on semantic information
processing. Infomaster integrates only structured information
sources, sources in which the syntactic form reflects its semantic
structure (in other words, databases and knowledge bases). This
restriction enables Infomaster to process the information in these
sources in a semantic fashion; information retrieval and
distribution can be conducted on the basis of content as well as
form.

The core of Infomaster is a facilitator that dynamically
determines an efficient way to answer the user's query using as
few sources as necessary and harmonizes the heterogeneities among
these sources. Infomaster connects, using wrappers, to a variety
of databases such as for example any SQL database through ODBC
and some World Wide Web sources. Infomaster contains a
model-elimination resolution theorem for abduction which acts as
a workhorse for the query planning process. The information
sources are described in terms of rules and constraints which are
stored in a knowledge base using Epilog, a main memory database
system. Informaster adopts mainly the local as view approach
where sources are described as views of the mediator relations.
It then employs an inversion mechanism to turn these into rules
for the mediator (or query) predicates and uses its abdcutive
theorem prover, together with constraint solving, to extract a
query plan from this. Infomaster has been in production use on
the Stanford campus since fall 1995 and is now commercially
available.

\item [InfoSleuth]

The InfoSleuth system architecture \cite{sleuth,sleuth2}
consists of a set of collaborating agents
that work together at the request of the user.
%
%
Users make requests to InfoSleuth from a domain-independent or
domain-specific applet. Requests are made against an ontology
specifying the users domain of interest. An ontology agent
together with a broker agent provide the basic support for
enabling the agents to interconnect and intercommunicate.
Ontology agents maintain a knowledge base of the different
ontologies used for specifying requests, and returns ontology
information as requested. The Broker agent maintains a knowledge
base of information that all the other agents advertise about
themselves, and uses this knowledge to match agents with
requested services. In this way the broker performs a form of
semantic matchmaking.

Within the InfoSleuth system, the agents themselves are organized
into layers, with the broker and ontology agents serving all of
the other agents. At the lower layers several other different
types of agents for processing information within InfoSleuth
exist: User agents, Resource agents, Task Execution agents and
Multiresource Query agents. The latter process complex queries
that span multiple resources. They may or may not allow the query
to include logically derived concepts as well as slots in the
ontology.

The Deductive Database system of LDL++ \cite{ldl++}
forms a component of the Infosleuth system. The
main function of  LDL++ in this system is implementing
the articulation axioms that support the mapping between
heterogeneous schemas. The advantages of LDL++  in this
context are enhanced by the LDL++ system's ability
of translating rules (including those with aggregates) into
equivalent SQL queries that are then offloaded for more
efficient execution to remote database servers \cite{ldl++}.

Infosleuth is used in applications for environmental data
exchange, analysis of genetic information and business
intelligence.

\item [MedLan]

The underlying basis of MedLan \cite{medlan} is the framework of
logic programming extended with: (i) the possibility of
partitioning the code into separate programs, (ii) the ability of
separate programs to interact, and (iii) the ability to answer
queries with respect to a combination of programs denoted by so
called {\em program expressions}.

The possibility of structuring logic programs into modules, that
may interact, allows one to implement classical mediator based
architectures, where the low-level modules can act as wrappers
for different data sources, while the intermediate level modules
act as mediators. The possibility of a dynamic combination of
programs by applying logic-based composition operators allows one
to construct mediators as semantic views on the data provided by
the wrappers. Among the operators for combining logic programs,
the {\em constrain} operator is of special importance for the
construction of mediators. It allows the use of a collection of
rules as a set of constraints over a logic program.

The capabilities of MedLan for constructing mediators have been experimented
in two application fields: semantic integration of deductive
databases and the construction of a declarative analysis level on top of a
traditional geographic information system.
Currently MedLan is also studied as a candidate for expressing security levels
for information systems.

\item [Ontobroker]
The Ontobroker-system \cite{ontobroker} is a WEB-based
application for ontology based search aimed at small communities
that are present in the internet or internet-like networks.
Ontobroker consists of a number of languages that allow us to
represent ontologies and to annotate web documents with
ontological information. It also contains a set of tools that
enhance query access and inference service in the WWW.
This tool set allows us to access information and knowledge from
the web and to infer new knowledge with an inference engine based
on techniques from logic programming. It aims to use semantic
information to guide the query answering process and to provide
information that is not directly represented as facts in the WWW
but which can be derived from other facts and some background
knowledge.

The Ontobroker  architecture consists of
three core elements: a query interface for formulating queries,
an inference engine used to derive
answers, and a webcrawler used to collect the required knowledge from the web.
Each of these elements is accompanied by a
formalization language: the query language for formulating queries, the representation language
for specifying ontologies, and the annotation language for annotating web documents with
ontological information.

The inference engine of Ontobroker is given a formal semantics
from Logic Programming. It uses generalized logic programs that
are translated further into normal logic programs via a
Lloyd-Topor transformation. Negation in the clause body that can
be non-stratified is interpreted via the well-founded model
semantics \cite{VanGelder93}. Standard techniques from deductive
databases, such as the bottom-up fixpoint evaluation procedure,
are also used in the implementation of Ontobroker.

Ontobroker has recently developed applications in the spirit of
the Semantic Web such as semantic community web portals and tools
for human resource knowledge management.

\item [PICSEL] \cite{picsel}
 is an information integration system where the mediator is based on the logical
 formalism of CARIN. 
This formalism combines the expressive powers of Horn rules and
description logics. CARIN is used as the core logical formalism to
represent both the domain of application and the contents of
information sources relevant to that domain. CARIN is a logical
formalism combining the expressive powers of function-free Horn
rules and description logics.

The strong use of a logical formalism allows PICSEL to have a
declarative definition of the relevant concepts for describing
the application domain and the information sources. This makes it
easy to take into account changes that can occur frequently e.g.
when new sources are added, old sources are removed, or when the
capabilities of existing sources are modified.
Also the formal semantics of PICSEL helps the designers to
express their knowledge in an unambiguous and rigorous way and to
define in a precise way the problem of answering queries w.r.t to
it.

In PICSEL the problem of answering queries can be identified as a
general problem of inference in a logical framework. Existing
well established techniques of this framework can help to
determine decidability and complexity results for the problem of
information integration, and to design correct and complete
algorithms. These algorithms have the advantage to be generic,
i.e. not specific to the application domain and to the sources and
therefore they can be reused in another application setting.

PICSEL has been tested with applications from the travel and
tourism domain. It is also used in electronic commerce
applications where PICSEL is employed as a tool for integrating
different services.

\end{description}

\subsection{Applications of Information Integration}

Although the area of Information Integration is relatively young
research in this field has had a strong emphasis on applications
from the very beginning of its existence. This emphasis on
practical aspects is growing with time. Much of the existing work
is grounded by the development of prototype systems and their use
for some specific application domain. Thus in many cases the
research is that of the development of methodologies for
information integration through principled engineering of
applications.

Applications of Information Integration can be divided into two
large classes: (a) integration of existing legacy heterogenenous
databases and (b) integration of information available over the
World Wide Web. At one end of the spectrum information
integration systems are developed for a particular domain of
interest. Examples include the Tambis \cite{Tambis} project
concerned with the problem of integrating Bioinformatics
Information Sources and the INEEL Data Integration System
\cite{ineel} that has been applied to problems of environmental
restoration. The main purpose of these applications is to offer
to the user an effective decision support tool through the
provision of extensive but relevant (to the users needs)
information. In such cases there is usually a rich amount of
domain specific knowledge that can be exploited in various ways
by the integration system e.g. in query optimization, providing
easy query formation and mediator specification, identification
and resolution of equivalences, etc.

At the other end of the application spectrum there is a fast
growing class of applications focusing on integration of Web data
resources. While Web-based integration systems usually provide
generic tools, particular applications focus on specific domains
of interest like entertainment \cite{theaterloc}, flight delay
prediction \cite{flight}, housing rentals (\cite{housing} and
Federal Goverment Information System \cite{federal}.
Moreover, information integration is an important enabling
technology for a wide class of electronic commerce applications
(see below for more details on this application domain). Issues
like rapid development of wrappers, flexible data models and
query languages that can easily accommodate semistructured data,
and efficient query execution, are crucial for such applications.

Another emerging new application domain is that of integration of
simulation results, so that we can also project information into
the future \cite{giofuture}. These type of applications will
complement existing applications, which only give a view into the
past and present, and hence address only one part of the needs of
decision-makers.

The web page of http://www-db.stanford.edu/LIC/companies.html
maintained by Gio Wiederhold lists 41 commercial suppliers of
Mediation Technology in the United States. Some of the more
commercial products include OmniConnect \cite{omni},
DataJoiner \cite{datajoiner},
Cross Access \cite{crossaccess}, and Enterworks \cite{enterworks}.

\subsection{Information Integration and E-commerce}

One important application where the need for information
integration would be ever growing is that of E-commerce over the
web. The number of people that that buy, sell of perform transaction on the Web
is increasing at a phenomenal rate.
Electronic commerce encompasses many issues, such as finding and filtering
information, securing information, generating dynamic supply-chain links,
online configuration of products and many others.

Many systems based on agent technology \cite{Maes2,Maes,Maes1}
are already present on the Web, {\em BargainFinder}
\cite{bargain} being the first agent for price comparison.
Systems of this category, often called {\em Shopbots}, can be
considered as a first attempt to link e-commerce with information
integration and agent technology. Clearly these systems perform
some form of integration of information at the different vendor
sources, but as the information they return is rather limited
(mainly price, avaliability, shipping time etc.) there is no need
for a sophisticated query translation algorithm as we have
described earlier. Other web-commerce agent systems like {\em
Kasbah} \cite{kasbah} pro-actively search for products that may be
of interest to the user, while {\em Firefly} \cite{firefly} is one
of the early systems that uses collaborative filtering to
recommend musical albums to its user. Advanced agent systems like
{\em AuctionBot} or {\em Kashab}, are capable of negotiating on
behalf of the user \cite{Maes2}. We therefore seem to be moving
towards a form of information integration that emphasizes the
aspect of personalization where the integration of information is
performed to suit each time the needs of the particular
user/buyer.

The functionality of current e-commerce agents is hindered by the
fact that information on the Web is currently in HTML format.
Agents use wrappers to extract, in a heuristic way, product or
other "content". Although there exist a number of approaches to
semi-automate this process, such ad-hoc solutions do not seem to
scale. Moreover, product information heterogeneity on the
semantic level, seems to be a more serious obstacle to efficient
business information exchange, than information representation
heterogeneity. In general, there is no agreement on fundamental
issues, such as what is included in a product domain, how to
describe products or how to structure product catalogs.

Product information heterogeneity can be tackled either by
standardization or integration \cite{Ng}. Industry realizing the
importance of resolving information heterogeneity has launched
several standardization initiatives. Some of these develop
horizontal standards (i.e. they cover all possible product areas)
such as UN/SPSC (Universal Standard Products and Services
Classification code, www.unspsc.org), while others develop
vertical standards (focusing on products of certain type) such as
RosettaNet for the area of hardware and software
products(www.rosettanet.org).

Two modern technologies, XML and Ontologies, are playing an
important role in these standardization efforts. XML has been put
forward as a important tool for tackling inter-operability
problems in e-commerce \cite{glushko}. Indeed, today there is a
growing number of XML standards for e-commerce capturing different
aspects of business activities \cite{Liu}. Although XML is a
major step forward, it should not be regarded as a solution to
all inter-operability problems, but more like a widely accepted
layer on which to build appropriate semantic information.
Although there is no single view on how to extend XML to support
greater inter-operability in e-commerce, it seems that ontologies
are becoming increasingly important as a component in
semantically rich e-commerce services, as advocated in
\cite{silver,smith,deborah}. For instance, buidling consensual
and reusable product catalogs is nothing more than building an
ontology for a certain domain. Indeed, there is a strong
interaction between the ontologies and online commerce
communities. Examples include Interprice \cite{Interprice} and
Content Europe \cite{ContentEurope} both providing ontology-based
support for e-commerce \cite{fensel-ieee}.


Although standardization efforts are important, integration
 still remains an issue, as
"most industrial standards are not quite mature at the current
stage, and there are no apparent leaders" \cite{Liu}. The
proliferation of standards threatens to create an electronic
marketplace dominated by "commerce islands", markets that have
become isolated by differing proprietary protocols and domain
standards. Therefore, it is inevitable that some of form of
integration will be required by the future e-commerce systems
\cite{Ng}. Logic could play an important role here in providing
higher levels of inter-operability needed for the integration of
information over different e-commerce markets.

Moreover, logics could contribute not only in solving product
information heterogeneity problems, but also to other aspects of
e-commerce. One such aspect is described in the recent work of
\cite{grosof99} that uses Logic Programming to develop a
framework for integrating business rules for electronic commerce.
These rules are expressed in a generalized form of logic
programs, called courteous logic programs, a framework that
incorporates a form of conflict handling. The declarative
semantics of this framework facilitates shared understanding and
inter-operability between different rules. Pilot applications in
e-commerce areas such as negotiations,  catalogs and storefronts
have been considered. The framework also supports an XML encoding
of courteous logic programs. A prototype system called {\em
CommonRules} is available at \cite{GrosofIBM}.

In addition, as the sophistication of e-commerce applications
increases, we also expect to see a stronger interaction between
the Semantic Web and E-commerce communities in the future. This
interaction could reveal new roles for logic in e-commerce
applications including the modeling of business rules
\cite{grosof99}, mentioned above, or the specification of workflow
\cite{bonner}.



\subsection{Logic Programming and the Web}

In the last five years there has been a new interest in the field
of Logic Programming aiming at linking Prolog languages to the
Web e.g. the PiLLow library for Internet/WWW Programming
\cite{Pillow}. The main idea is to provide a facility so that
pages can be downloaded from the Web and turned into a
corresponding Prolog program containing information extracted from
these web pages. In general, this is an attempt to encode
information pertaining to a web page, which can be either
information within the page or meta information about the web
page, into the declarative form of a Prolog program.

One can then view this as a mechanism for information integration
under the paradigm of mediation as these logic programs define a
common mediator layer for the various web pages retrieved from the
web in this way. Analyzing and synthesizing the information in
these logic programs constitutes a primitive form of information
integration over the web page sources. PrologCrawler
\cite{PrologCrawler} and ExpertFinder \cite{ExpertFinder} are two
such systems that use this "web page to prolog" approach in order
to integrate information from various web pages. Similarly,
WebLog \cite{weblog} and $D^{3}$Web \cite{D3web} are Datalog based
query languages for information held over several web pages.

The LogicWeb system \cite{LogicWeb,LogicWeb2} converts web pages
and their links into logic programs with the additional
possibility for a web page itself to contain LogicWeb code. These
logic programs can then be semantically composed together in
several ways thus achieving the integration of information from
the original web page sources. Applications of LogicWeb include a
citation search tool \cite{LogicWebTool} and a system for
web-based guided tours \cite{LogicWebTours}.

These ideas to convert information from an HTML document of a web
page to a logic program and utilize these programs to
declaratively synthesize the information, extend from HTML web
pages to XML documents \cite{datalogforXML} and RDF descriptions
\cite{que-inf-rdf} thus enabling a higher-level of semantic form
of information integration.

\section{Challenges of CL in Information Integration}

The problem of information integration has been the subject of
intensive research activity over the recent years, with  most of
the work concentrating on modeling data sources in a single
unified view. While significant progress has been made, many
important problems remain unsolved. Future developments in
information integration are expected to center around the
following inter-related research themes as presented below. In the
discussion of these problems we will focus particularly on the
potential role that logic could play in addressing them. Some of
these links to logic reflect the authors'
personal opinion.\\[2pt]

\noindent {\bf Representation/Optimization}: As noted earlier most
of the early work on information integration
 has been carried out under the assumption
that information sources can be modeled as relational databases.
Logic based approaches have followed to a large extend this line
of research. However, nowdays interest is shifting fast towards
semistructured data. Modeling, storing, managing and querying
such data are emerging as important problems and will receive
much attention in the years to come. Moreover, the development of
standards for data representation and exchange on the Web, like
XML, have already a strong impact on data modeling for information
integration. The similarity between XML and data models like OEM
developed independently by the DB community, will further
facilitate the study of problems related to XML data management
\cite{widomdirect}. The role of CL in modeling semi-structured
data is emerging as an important question that has received
relatively little attention so far.


On the other hand, it is now clear that information integration
in a dynamic environment like the internet, will need to employ
effective query optimization and execution techniques.  Richer
forms of knowledge about the content and the structure of the
sources as well as their inter-relations, are likely to be
important elements in the design of future information
integration systems facilitating new forms of semantic query
optimization. Inductive learning and data mining techniques can
further assist in the extraction of semantic knowledge for query
optimization. Query optimization can also be enhanced within a
logical framework with the further use of integrity constraints
by interleaving their satisfaction with the query reformulation
process. Integrity constraints can express information about
completeness, preferences, inclusion and other properties of data
sources.

The main focus of the modeling methods for information integration
is on providing a mediated schema that defines the ``semantics''
of the underlying sources in the strict context of the mediated
service that the global schema supports. Therefore, there is no
need to define the intended meaning of the objects that make up
the mediated schema. However, large scale information
integration, as conceived for instance by the Semantic Web
initiative,  requires tackling data representation problems in a
more global context, because "...we expect this data, while
limited within an application, to be combined later with data
from other applications" \cite{semweb}.

Integration in the large calls for a semantically rich
representation of the data that supports sharing, re-usability
and extensibility.
These requirements render metadata and ontologies, as key
components of information integration systems
\cite{hull,ouksel,sheth1,wie94}. Metadata provides information
about data, eg. information about database schemas and their
intended meaning, which can be captured in an ontology that
defines a common ``vocabulary'' for describing different
information sources and services.
The intended meaning of a database schema will be specified
in terms of an ontology, and from this ontology  (possibly
together with some inter-ontology mappings) the mediator will be
able to specify, automatically and through reasoning, the
relationship between this schema and other schemata defined with
respect to the same or related ontologies. Similarly, the user
can use the same or related ontology to pose queries to the
sources.

Several information integration systems including Information
Manifold \cite{lreview}, OBSERVER \cite{observ99}, Ontobroker
\cite{ontobroker}, InfoSleuth \cite{sleuth,sleuth2} and SHOE
\cite{shoe} use ontologies strongly in their representation
language. As noted earlier, ontologies also play a crucial role in
the Semantic Web architecture. Logic, and in particular
description logics, form an integral part of ontologies as most of
these systems use them as their basic formalism for implementing
ontologies.
Formulating ontological context in logic (as in \cite{fikes1}) and
reasoning about properties of ontologies using non-monotonic
reasoning methods (e.g. default or hierarchical reasoning) seems
a useful direction of research.

It is also expected that there will be a growing need for more
advanced forms of application layers that would enable
specialized and advanced forms of integration  as compared with
the relatively shallow integration that is carried out by today큦
systems. This will require declarative and more expressive
mediator and application layers providing more flexible data
models. Again the use of logic is one alternative for this
purpose. For example, explicit negation in Logic Programming and
constraints as in Constraint Logic Programming can be used to
enrich the representation language. Current experience suggests
that this use of logic will also need the use of other data
models and computational methods in order to enhance the
computational effectiveness of the overall framework. In such a
hybrid model, logical reasoning can provide one of the main
channels of cooperation between the different computational
processes involved.

The work on Semantic Web services \cite{DAML2} is characteristic
of the trends in the design of the next generation of advanced Web
applications. Its main purpose is to provide declarative
Application Program Interfaces that will enable apart from service
discovery, additional features such as automated service
execution and more automated service composition and
inter-operation. The solution proposed in \cite{DAML2} involves a
combination of semantic markup of Web pages using DAML languages
together with an agent infrastructure that uses situation
calculus and the ConGolog agent programming language
\cite{ConGolog}.


Two important technical problems that need to be addressed further
irrespective of the particular form of more expressive language
representation that we use are the problems of {\em incomplete
information} and {\em semantic conflicts or contradictory data}.
The problem of incompleteness can appear either at the level of
the data sources themselves where some information is missing or
at the level of the description of the mediator architecture as
for example with semi-structured data. In the later case one
issue to address is how we can use logic to describe the partial
or uncertain information and then use mechanisms that are capable
of reasoning under such incomplete information. The problem of
incompleteness of data sources in information integration was
studied in \cite{etzioni} and shown to be related to
non-monotonic reasoning. As such the logical techniques of
negation by failure, default reasoning and abduction could be
useful in addressing this problem. In particular, constructive
abduction that allows non-ground hypotheses seems to be well
suited for query reduction under incomplete information. This can
give existential answers conditional on an set of associated
constraints expressing the range of values that a missing data
entry can take.

It is inevitable that semantic conflicts or contradictory data
will appear in information integration particularly when this is
done over disparate sources over the WWW where there is no
central control on the data available in these sources. The
simplest form of this problem is that of naming mismatch where
different syntactic names are given to the same semantic entity.
WHIRL \cite{whirl} combines techniques from information retrieval
and AI to address this problem through an appropriate similarity
measure. In this way it provides a system that is able to reason
approximately (but with levels of confidence) with the partial
information that it extracts from the unstructured textual form
of the information sources.

These two inter-related problems of incomplete information and
conflict resolution have been at the heart of many studies in
Computational Logic, e.g. in default reasoning, extended logic
programming, preference reasoning and argumentation. We can then
examine ways how these largely theoretical methods can be adapted
and used in the more practical setting for applications of
Information Integration.\\[2pt]

%

\noindent {\bf Automation}: For information integration systems
to scale up we will need to automate at least some part of their
development.  Currently, the task of the description of the data
sources that are available to a system is undertaken by the
creator of the information integration system.
But, when the number of sources grows, hand-coding the mapping
between the mediated schema and the sources, is a major
bottleneck in deploying large-scale integration systems.
Therefore, there is a great need for methods and tools that
assist or automate the process of generating source descriptions.


There are several places where machine learning can help in the
automation of information integration. These include (i) the
extraction of information from sources as for example in the
process of automatic wrapper induction
 \cite{eikvil,muslea-intro,kushmerik,craigcia,Soderland97},
(ii) learning mappings between mediator relations and source
schemas thus automatically constructing part of the central
mediator architecture
\cite{PE95,levy-source-learn,milo,calabria-scheme-1,calabria-scheme-2}
and (iii) extracting (additional) regularities over the data
sources and mediator relations as meta-level integrity
constraints that would be useful in the process of query planning
and optimization. In many cases, these learning tasks take the
usual concept learning paradigm of learning a set of concepts from
given training examples e.g. learning concepts in the mediator
schema from labeled sources of information as instances of these
relations. This type of learning has been extensively studied in
the field of Inductive Logic programming (ILP) and its methods
have already began to be used in these problems
\cite{Freitag98,Craven-et-al,Junker}. The significance and
potential of computational logic for these problems is high.
Indeed, the inherent relational nature of any Information
Integration framework makes the relational learning framework of
ILP particularly appropriate for the aforementioned machine
learning tasks within the general task of Information Integration.

As discussed in the previous section, the use of ontologies is an
important approach to defining large-scale mediation services.
They support sharing, reusability and extensibility, which are
all important features for the rapid development of information
integration applications. It is therefore natural to expect that
the different aspects of the problem of learning ontologies, as
studied in \cite{ieee-learn,embley,ecaiws,ijcaiws} and
\cite{borys} where a review of existing approaches is given, will
become important in the future. Relational learning can play a
central role in this effort to automate the construction of the
rich structure of ontology based mediators.\\[2pt]

\noindent {\bf Personalization}: In a mediator-based information
integration system two interfaces have to be implemented, the
mediator/source interface and the user-application/mediator
interface. Indeed, the overall problem of information integration
can be split into two levels:

\begin{itemize}
\item Interpret what the user (or application program) is asking for in
her query (or high-level goal).
\item Answer the query (or goal)  through a suitable integration of information
of various sources available.
\end{itemize}

While much work has been done at the second level based on a
mediator architecture that describes sources, i.e. the
mediator/source interface, the problem of modeling
users/application for information integration is less well
understood. This first level involves understanding the needs of
each individual in a given context and satisfying these needs in
the best possible way. This is a complicated problem with many
different aspects. Existing user agent systems  like {\em
Webwatcher} \cite{webwatch} and {\em Letizia} \cite{letizia}
assist the user in locating information on the Internet, by
tracking the user큦 behaviour. On the information provider side, a
current popular approach to personalization is the use of data
mining techniques on Web-logs that track the user큦 browsing
behaviour represented as a clickstream
\cite{aggarwal,srivastava,ieee-mine}.
The discovered patterns can be used in
different ways such as changing the web structure for easier
browsing, predicting future page requests, predicting user
preferences for active advertising, or making recommendations to
the user.

However, the above approaches either rely heavily on user browsing,
or are restricted to individual web sites and therefore they do not
adequately address all aspects and forms of information integration.
The problem of personalization in a general information
integration setting,
involves understanding the user (or application) needs  in a way
that will allow the computation of the most appropriate queries to
the sources that would satisfy these needs. In other words, a {\em
high-level query formation} is required that is sensitive to the
particular user큦 context and needs. In approaches where the user
formulates her queries directly in the mediator language, it is
assumed that she is familiar with the vocabulary available for
posing queries, or with the range of information that is
available to the mediator \cite{fikes1}. In application domains
where this is not the case, the user should be assisted in
formulating her queries \cite{fikes1,interact}.

Moreover, answering user queries in a satisfactory way, involves
in many cases a level of understanding that can not be reached
without some semantic knowledge about the data and the context of
the query. Although query answering can be seen as a process of
matching the query description with the source description, in
many interesting cases a simple syntactic matching is not
adequate and semantic information is required. Ontology based
query formulation is a first step towards providing some of the
necessary functionality and has been employed in some of the
existing systems \cite{sheth1,observ99,ontobroker}. On the other
hand, we expect future work on personalization to be more tightly
linked to the emerging metadata languages RDF and RDFS
\cite{metu}. In addition, current approaches to personalization
such as collaborative filtering or user clustering, will evolve to
accommodate semantically richer forms of information that will
become available on the internet.


In general, effective user profiling \cite{pohl} and information
brokering based on this, is a task that depends heavily on the
existence and use of extensive background knowledge about the
application domain and requires advanced techniques of knowledge
representation and reasoning. "If we want our computers to understand
us, we will need to equip them with adequate knowledge" \cite{minsky}.
Methods from ILP for knowledge
intensive learning (e.g. \cite{parsons98}) and the use of CL for
advanced forms of reasoning, such as default
reasoning with (user) preferences and constraints, could prove
useful in this task, especially in the context of the future
Semantic Web services.\\[2pt]

\noindent
{\bf Future Forms of Mediation}:
%
Future developments will see an ever increasing number of
applications that use information integration over the Web and in
particular the emerging Semantic Web. This would range from more
advanced search engines to applications for specific domains where
integration of information relative to a user's general needs is
performed pro-actively. The two phases of query formation and
query planning would then be strongly interleaved for this
purpose.

As we automate more and more the construction process of a
mediator this evolves into a {\em facilitator} as defined in
\cite{wie97}. A facilitator is dynamic and responsive to changing
situation. It will be able to accept in a dynamic way meta-data
about information sources and logical statements relating
disparate concepts of an underlying ontology in order to
automatically integrate a new resource into the system. There
will also be an increasing need for {\em abstraction} where the
volume of synthesized data is reduced while maintaining its
essential (for the application) information content. For example,
instead of responding to a query with a set of all answers we can
use a rule or intentional answer that characterizes all the
extensional answers. Clearly, logic can help realize these
characteristics features of future mediators.

\subsection{Multi Agent Systems for Information Integration}

Information integration on the Web presents us with different
challenges that require scalable, flexible and extensible
solutions. Recent developments in agent systems seem to provide a
promising supporting technology for realizing large scale
information integration solutions. The Infosleuth approach
\cite{sleuth,sleuth2}, and its predecessor Carnot project,
pioneered  the use of agent technology in information integration.
In these systems CL had a significant role to play. The broker
agent, who is responsible for pairing agents seeking for a
service with agents that can perform that service, is partially
implemented in the logical deductive Database language LDL++
\cite{ldl++}. More recently, the potential role of CL-based
agents in information integration has been demonstrated by a
number of works e.g. \cite{martelli,xanthakos}.

The CL-agent approach brings together the benefits of declarative
specification and rich level of expresiveness offered by
computational logic, with a number of other benefits derived from
the agent architecture. These additional benefits include:
Reactivity: alertness to changes in the user requirements, and
changes to the location, availability and content of information
sources; Interactivity of the systems with the user; and
Interleaving of query planning and query plan execution: the
mediator can liaise with the information sources while it is
constructing a query plan, and before a complete plan is
constructed. This has the advantage of pruning the search for a
plan as early as possible. For example, it can alert the planning
process of the unavailability of an information source, or it can
find values for partial queries that will substantially reduce
further search.

The approach of \cite{xanthakos}, uses the agent architecture of
Kowalski and Sadri together with techniques of abduction. It
adopts the global as view approach, where global relations are
defined in terms of the information stored at the sources and
incorporates the expression of functional dependencies which can
be utilized in query planning. Furthermore it allows the
expression and use of priorities amongst information sources in
terms of  their reliability and degree of completeness with
respect to items of data.

We expect that in the near future, as applications become more
complex, a closer link between information integration and agent
technologies will be established. Future information integration
systems will require agents for different activities, over and
above those for the mediator, such as those for user profiling,
symbolic learning and provision of user-friendly interfaces. The
use of logic-based agents for information integration paves the
way to the synthesis of information integration with other
techniques to give such more powerful systems.

\section{Conclusions}

The area of Information Integration is a young but fast growing
field that has emerged from the need to exploit more fully the
available data spread over various databases of different type.
It has been given a special impetus with the appearance of the
World Wide Web where the need for new research on the problem of
integration of information spread over the web is considered to
be of paramount importance not least because of its enormous
potential for commercial exploitation. This development has meant
that together with new database techniques there is an ever
growing need for the use of AI techniques such as those of
multi-agent systems, knowledge representation (including in
particular ontologies and hierarchies), natural language,
resolution of conflict and machine learning. In fact, turning
this around, the problem of information integration over the Web
is providing an excellent opportunity for a new experimental
arena where AI theory and techniques can be applied and tested.

To the extent that Computational Logic belongs (also) to AI (see
the recent book on "Logic-Based Artificial Intelligence" published
after a meeting held in Washington in June 1999 \cite{LB-AI-99})
we can see that Logic will also have a role to play in this new
research area. A central role of logic in information integration
concerns the problem of specifying a mediator. This is analogous
to the problem of views in databases and, as exposed early in
\cite{ullman97}, it is possible to use logic to formalize most if
not all mediator architectures that have been proposed. In some of
these approaches logic is used explicitly to specify and to a
certain extent implement the mediator architecture.

Indeed, the potential usefulness of logic was realized from the
very beginning of investigations in this problem. As the work
developed to address the problem in a more complete way the role
of logic was exposed more clearly to be that of a facilitator for
the cooperation between the different other computational
processes involved in an information integration system. Logical
inference can be used as the mechanism for communication and
intelligent cooperation between these processes. We expect that
this central role of logic will be further exposed as the
ontologies used for mediation get richer and there is more scope
for reasoning with these ontologies. This would help to develop
more advanced forms of integration compared with the relatively
shallow integration of today's frameworks and systems. Also logic
can be used to link the advanced and specialized (domain
specific) needs of the application to the more general (problem
independent) lower layers of the mediator architecture. An
alternative way to formalize this upper application layer is
using an algebra (see the SKC project \cite{SKC-project}) and
hence the merits of each approach, logical or algebraic, need to
be compared.

A specific problem of immediate importance is that of
rationalizing semi-structure data in the same way that logic
rationalizes databases. This concerns the use of logic to
formalize directed graphs of semi-structured data in the same way
that we formalize in logic the (relational) tables of databases
and their query language. Work in this direction has already
started \cite{bertram,lenzerini1}, opening a new opportunity for
logic-based databases. More generally, logic can help as a
unifying basis with which we can add structure to the meaningful
content of web pages so that a higher-level of semantic
information integration can be performed over the Web. Linking
logical inference with ontological information is an important
next development for building this vision of the Semantic Web.

Comparing the potential  role of logic in Information Integration
with its role in other areas in the past e.g. in the development
of Constraint Programming, we see again that when we consider an
application domain in its entirety the central role of logic is
to provide a modeling environment for the problem (in our case
the overall mediator architecture) and the link of this
description to other computational methods needed to solve the
problem.
Of course, if and when some of these other computational
processes can themselves be performed in a logical setting this
central communicator role of logic can be implemented more
tightly enabling more functionality. In the problem of information
integration such cases are  the use of (i) Inductive Logic
Programming for the automatic generation of mediators, (ii)
logic-based multi-agents as a framework for implementing the
overall communication layers of a mediator architecture and (iii)
Constraint Logic Programming to help address the scaling problem
through the use of meta-level constraint information in query
planning.

We would like to end this survey with a quote from a recent
article \cite{SemWebBernels} of Tim Bernels-Lee, James Hender and
Ora Lassida: "Adding logic to the Web ... is the task before the
Semantic Web community at the moment". This statement reveals
succinctly the potential role of computational logic in the
present and future development of Information Integration.



\vspace*{.4cm} \noindent {\bf Acknowledgements} This paper is a
follow up of a survey on the subject of Information Integration
carried out by Compulog-Net within its wider effort to identify
future promising technological developments for Computational
Logic. Mimmo Sacca, the coordinator of Compulog-Net for the area
Logic-based Databases, has helped in collecting material and in
forming the structure of this survey. We would also like to thank
the following colleagues who provided us with useful material and
their views on the subject by answering a questionnaire that was
sent to them: Stephanne Bressan, Stefano Ceri, Thomas Eiter,
Wolfgang May, Maurizio Lenzerini, Alon Levy, Yannis
Papakonstantinou, Dino Pedreschi, Marie-Christine Rousset, Fariba
Sadri, V.S. Subrahmanian, Letizia Tanca, Francesca Toni, Franco
Turini, Jeff Ullman,  Gio Wiederhold and Carlo Zaniolo. Their
replies have been integrated at appropriate places in the paper.

\begin{small}

\end{small}


\begin{thebibliography}{99}

\bibitem{abiteboul}
S. Abiteboul.
Queryring semistructured data.
{\em ICDT 97}.

\bibitem{abibook}
S, Abiteboul, P. Buneman, D. Suciu
{\em Data on the Web : From Relations to Semistructured Data and XML}.
Academic Press/Morgan Kaufmann, 1999.

\bibitem{ad}
S. Abiteboul, O. Duschka.
Complexity of Answering Queries Using Materialized Views.
{\em ACM Symp. on the Theory of Computing}, 1998.

\bibitem{lorel1}
S. Abiteboul, D. Quass, J. McHugh, J. Widom, J. Wiener.
The Loerl Query Language for Semistructured Data.
{\em International Journal on Digital Libraries}, 1, 1997.

\bibitem{aggarwal}
C. Aggarwal, P. Yu.
Data Mining Techniques for Personilization.
{\em IEEE Data Engineering Bulletin}, 23(1), 2000.

\bibitem{medlan}
D. Aquilino, P. Asirelli, C. Renso, F. Turini.
Using MedLan to Integrate Geographical Data
{\em Journal of Logic Programming}, to appear.

\bibitem{araneus}
ARANEUS Project. http://www.dia.uniroma3.it/Araneus/index.html

\bibitem{darpa}
Y. Arens, R. Hull, R. King et al.
Reference Architecture for the Intelligent Information Integration. Version 2.0, 1995.
http://www.darpa.mil/iso/i3/

\bibitem{sims}
Y. Arens, C. Knoblock, W.M. Shen.
Query Reformulation for Dynamic Information Integration.
{\em Journal of Intelligent Information Systems}, 6(2/3), 1996.

\bibitem{ecp97}
N. Ashish, C. Knoblock, A. Levy.
Information Gathering Palns with Sensing Actions.
{\em ECP'97}.

\bibitem{Tambis}
P.G. Baker, A. Brass, S. Bechhofer, C. Goble, N. Paton, R. Stevens. TAMBIS:
 Transparent Access to Multiple Bioinformatics Information Sources. An Overview.
{\em Proc. of the Sixth Inter. Conf. on Intelligent Systems for Molecular
Biology, ISMB98}, 1998.

\bibitem{theaterloc}
G. Barish, C. Knoblock, Y. Chen, S. Minton, A. Philpot, and C. Shahabi.
TheaterLoc: A Case Study in Information Integration.
{\em Proc. IJCAI Workshop on Intelligent Information Integration}, 1999.

\bibitem{sleuth}
Bayardo et al.
Semantic Integration of Information in Open and Dynamic Environments.
{\em SIGMOD'97}.

\bibitem{beeri}
C. Beeri, A. Levy, M.C. Rousset.
Rewritting queries using views in description logics.
{\em PODS'97}.

\bibitem{momis}
S. Bergamaschi, S. Castano and M. Vincini.
Semantic Integration of Semistructured and Structured Data Sources.
{\em SIGMOD Record}, 28(1), 1999

\bibitem{SW-roadmap}
T. Berners-Lee.
Semantic web road map.
http://www.w3.org/DesignIssues/Semantic.html, September 1998.

\bibitem{semweb}
T. Berners-Lee, D. Connoly, R. Swick.
Web Architecture: Desrcibing and Exchanging Data.
http://www.w3.org/1999/04/WebData, 1999.

\bibitem{SemWebBernels}
T. Bernels-Lee, J. Hender, O. Lassida. The Semantic Web.
Scientific American, May 2001.

\bibitem{hu}
E.Bertino, B.Catania, V. Gervasi, A Raffaeta.
A Logic Approach to Cooperative Information Systems.
{\em Journal of Logic Programming}, to appear.

\bibitem{boley}
H. Boley.
Relationship Between Logic Programming and RDF.
{\em Pacific Rim International Workshop on
Intelligent Information Agents, PRIIA 2000}.

\bibitem{PrologCrawler}
E. Bolognese, A. Brogi.  A  Prolog Meta-Search Engine for the
World Wide Web. Techical Report, University of Pisa, Italy, 1999.

\bibitem{XQL-compare}
A. Bonifati, S. Ceri.
Comparative analysis of five XML query languages.
{\em SIGMOD Record}, Vol 29, 2000.

\bibitem{bonner}
A. Bonner. Workflow, Transactions and Datalog. {\em PODS'99}.

\bibitem{multiDB2}
Y. Breitbart, A. Silberschatz. Multidatabase Update Issues. {\em
Proceedings of SIGMOD 1988}, pp. 135-142, 1988.


\bibitem{coin1}
S. Bressan, K. Fynn, C. Goh, S. Madnick, T. Pena, M. Siegel.
Overview of the Prolog Implementation of the COntext INterchange
Prototype. {\em Fifth International Conference on Practical Applications of Prolog},
1997.

\bibitem{coin2}
S. Bressan, C. Goh.
Answering Queries in Context.
{\em International Conference on Flexible Query Answering},1998.

\bibitem{coin3}
S. Bressan, C. Goh.
Semantic Integration of Disparate Information Sources over the Internet
Using Constraints.
{\em Constraint Programming Workshop on Constraints and the Internet},
1997.

\bibitem{RDFS}
D. Brickley, R.V. Guha.
Resource Description Framework (RDF) Schema Specification 1.0.
http://www.w3.org/TR/2000/CR-rdf-schema-20000327, 2000.

\bibitem{web-formal}
J. Broekstra, M. Klein, S. Decker, D. Fensel, I. Horrocks.
Adding formal semantics to the Web: building on top of RDF Schema.
{\em Proc. of the "ECDL 2000 Workshop on the Semantic Web''},
http://www.ics.forth.gr/proj/isst/SemWeb/program.html.

\bibitem{ExpertFinder}
A. Brogi, G. Marongiu. ExpertFinder: A Prolog Recommender System
Integrated with the WWW. {\em Proc. of the 1999 Joint Conf. on
Declarative  Programming AGP-99}, 1999.

\bibitem{buneman}
P. Buneman. Semistructured data. {\em PODS 97}.

\bibitem{unql}
P. Buneman, S. Davidson, G. Hillebrandt, D. Suciu.
A Query Language and Optimization Techniques for Unstructured Data.
{\em SIGMOD 96}.

\bibitem{bunemanstr}
P. Buneman, S. Davidson, G. Hillebrandt, D. Suciu.
Adding structure to unstructured data.
{\em ICDT'97}.

\bibitem{Pillow}
D. Cabeza, M. Hermenegildo, S. Varma. The PiLLoW/CIAO Library for
INTERNET/WWW Programming. {\em Proc. Of the 1st Workshop on Logic
Programming Tools for Internet Applications}, P.Tarau, A.Davison,
K.De Bosschere, and M.Hermenegildo (eds.), JICSLP-96, pp. 43-62,
1996.

\bibitem{lenzerini1}
D. Calvanese, G. De Giacomo, M. Lenzerini.
What can Knowledge Representation do for Semistructured Data?
{\em AAAI'98}.

\bibitem{lenzerini2}
D. Calvanese, G. De Giacomo, M. Lenzerini.
Semi-structured Data with Constraints and Incomplete Information.
{\em Proc. of the 1998 Description Logic Workshop, DL'98}.

\bibitem{lenzerini3}
D. Calvanese, G. De Giacomo, M. Lenzerini, D. Nardi, R. Rosati.
Description logic framework for information integration.
{\em KR'98}.

\bibitem{lenzerini4}
D. Calvanese, G. De Giacomo, M. Lenzerini.
Answering queries using views over description logics knowledge bases.
{\em AAAI'00}.

\bibitem{lenzerini5}
D. Calvanese, G. De Giacomo, M. Lenzerini, D. Nardi, R. Rosati.
Information integration: Conceptual modeling and reasoning support.
{\em CoopIS'98}.

\bibitem{CGLV99}
D. Calvanese, G. De Giacomo, M. Lenzerini, M. Vardi.
Rewriting of regular expressions and regular path queries.
{\em PODS'99}.

\bibitem{CGLV00-1}
D. Calvanese, G. De Giacomo, M. Lenzerini, M. Vardi.
Answering Regular Path Queries Using Views.
{\em ICDE'00}.

\bibitem{CGLV00-2}
D. Calvanese, G. De Giacomo, M. Lenzerini, M. Vardi.
View-Based Query Processing for Regular Path Queries with Inverse.
{\em PODS'00}.

\bibitem{catarci}
T. Catarci, M. Lenzerini.
Representing and using interschema knowledge in cooperative information systems.
{\em Journal of Intelligent and Cooperative Information Systems}, 1993.

\bibitem{kasbah}
A. Chavez, P. Maes.
Kasbah: An Agent Marketplace for Buying and Selling Goods.
{\em Proc. of the 1st
Inter. Conf. on the Practical Appication of Intelligent Agents and Multi-Agent Technology},
London, UK, April 1996.

\bibitem{XML-challenge}
S. Ceri, P. Fraternali, S. Paraboschi.
XML: Current Developments and Future Challenges for the Database Community.
{\em EDBT 2000}.

\bibitem{XML-GL}
S. Ceri, S. Comai, E. Damaini, P. Fraternali, S. Paraboshi, L. Tanca.
XML-GL: A graphical language for querying and restructuring XML documents.
{\em Proc. of the 8th WWW Conference}, 1999.

\bibitem{dataware1}
S. Chaudhuri, U. Dayal. An overview of Data Warehousing and OLAP
technology, {\em ACM SIGMOD Record}, Vol. 26(1), pp. 65-74, 1997.

\bibitem{metu}
I. Cingil, A. Dogac, A. Azgin.
A Broader Approach to Personalization.
{\em Communications of the ACM}, 43(8), 136-141, 2000.

\bibitem{whirl}
W.W. Cohen. Integration of heterogeneous databases without common domains
using queries  based on textual similarity. {\em SIGMOD'98}.

\bibitem{review}
W. Cohen, C. Knoblock, A. Levy, S. Minton.
Trends and Controversies: Information Integration.
{\em IEEE Intelligent Systems Journal}.

\bibitem{log-inter}
W. Conen, R. Klapsing.
A Logical Interpretation of RDF.
{\em Electronic Transaction on AI. Semantic Web Area}.
http://www.etaij.org/seweb/.

\bibitem{CGRDF}
O. Corby, R. Dieng, C. Hebert.
A Conceptual Graph Model foe W3C Resource Description Framework.
To appear in {\em Proc. of the 8th International Conference on Conceptual Structures
ICCS'2000}, LNCS 1867.


\bibitem{Craven-et-al}
M. Craven, D. DiPasquo, D. Freitag, A. McCallum, T. Mitchell, K. Nigam,
S. Slattery. Learning to extract sympolic knowledge from the world wide web.
{\em AAAI'98}.


\bibitem{martelli}
P. Dart, E. Kazmierczak, M. Martelli, V. Mascardi, L. Stirling, V.S.
Subrahmanian, F. Zini. Combining logical agents with rapid prototyping for engineering
distributed applications. {\em submitted}.

\bibitem{LogicWeb2}
A. Davison. A Concurrent Logic Programming Model of the Web. {\em
Proc. of the 1st Int.Conf. on Constraint Technologies and Logic
Programming, PACLP-99}, pp. 437-451, 1999.

\bibitem{multiDB3}
U. Dayal. Processing Queries Over Generalization Hierarchies in a
Multidatabase System. {\em Proceddings of VLDB 1983}, pp.
342-353, 1983.

\bibitem{que-inf-rdf}
S. Decker, D. Brickley, J. Saarela, J. Angele. A Query and
Inference Service for RDF. {\em W3C Query Languages Workshop},
1998.

\bibitem{ConGolog}
G. De Giacomo, Y. Lesperance, H. Levesque.
ConGolog, a Concurrent Programming Language based on
Situation Calculus.
{\em Artificial Intelligence}, 121(1-2), p. 109-169, 2000.

\bibitem{LucBook96}
L. De Raedt. {\em Advances in Inductive Logic Programming}, IOS Press, 1996.

\bibitem{STORED}
A. Deutsch, M. Fernandez, D. Suciu.
Storing Semistructured Data with STORED.
{\em SIGMOD'99}.

\bibitem{XML-QL}
A. Deutsch, M. Fernandez, D. Florescu, A. Levy, D. Suciu.
XML-QL: A query language for XML.
{\em W3C Workshop on Query Languages}, 1998.


\bibitem{levy-source-learn}
A. Doan, P. Domingos, A. Levy.
Learning Source Descriptions for Data Integration.
{\em Inter. Workshop on The Web and Databases (WebDB'00)}, 2000.


\bibitem{duschka97}
O. Duschka.
Query Optimization Using Local Completeness. {\em AAAI 97}.

\bibitem{info2}
O. Duschka, M. Genesereth. Query Planning in Infomaster.
{\em SAC '97}.

\bibitem{dg}
O. Duschka, M. Genesereth.
Answering Recursive Queries Using Views.{\em PODS 97}.

\bibitem{dl}
O. Duschka, A. Levy.
Recursive Plans for Information Gathering. {\em IJCAI 97}.

\bibitem{eclipse}
ECLiPSe User Manual.
ECRC, Munich, Germany, 1994.

\bibitem{eikvil}
L. Eikvil.
Information Extraction from World Wide Web.
{\em Report 945}, Norwegian Computing Center, 1999,
available from http://www.nr.no/~eikvil/online.html.

\bibitem{embley}
D. Embley, D. Campbell, S. Liddle, R. Smith.
Ontology-Based Extraction and Structuring of Information from
Data-Rich Unstructured Documents.
{\em CIKM'98} ,1998.

\bibitem{etzioni}
O. Etzioni, K.Golden, D. Weld.
Tractable closed world reasoning with updates.
{\em Artificial Intelligence},89, 1996.


\bibitem{silver}
D. Fensel.
{\em Ontologies: Silver Bullet for Knowledge Management
and Electronic Commerce}, Springer, forthcoming.

\bibitem{IEEE1}
D. Fensel (editor).
The semantic Web and its languages.
{\em IEEE Intelligent Systems},  Vol. 15(6), p. 67-73, 2001.

\bibitem{fensel-ieee}
D. Fensel.
Ontologies and Electronic Commerce.
{\em IEEE Intelligent Systems},  Vol. 16(1), p. 8, 2001.

\bibitem{ontobroker}
D. Fensel, S. Decker, M. Erdmann, R. Studer. Ontobroker:
The Very High Idea. In {\em 11th International FLAIRS Conference,
FLAIRS-98}.

\bibitem{OIL1}
D. Fensel, F. van Harmelen, I. Horrocks, D. McGuinness,
P. Patel-Schneider.
OIL: An Ontology Infrastructure for the Semantic Web.
{\em IEEE Intelligent Systems},  Vol. 16(2), p. 38-45, 2001.

\bibitem{IEEE2}
D. Fensel, M. Musen (editors).
The Semantic Web.
{\em IEEE Intelligent Systems},  Vol. 16(2), p. 24-79, 2001.

\bibitem{fikes1}
R. Fikes, A. Farquhar, W. Pratt.
Information Brokers: Gathering Information from Heterogeneous Information
Sources. Eckerd College, Key West, Florida, 1996.
http://ksl-web.stanford.edu/KSL\_Abstracts/KSL-96-18.html.

\bibitem{calabria-ssd-2}
S. Flesca, S. Greco.
Partially Ordered Regular Languages for Graph Queries.
{\em Proc. of The 26th International Colloquium on Automata, Languages
andProgramming (ICALP)}, Praga, 1999.

\bibitem{calabria-views-1}
 S. Flesca, S. Greco.
 Rewriting Queries Using Views.
 {\em Proc. of 10th
International Conference on Database and Expert System Application
 (DEXA'99)} , Firenze, 1999.

\bibitem{calabria-ssd-1}
S. Flesca, S. Greco.
Querying Graph Databases.
{\em Proc. Conference on Extending Database Technology (EDBT'00)},
Konstanz, 2000.

\bibitem{calabria-fddbs}
S. Flesca, L. Palopoli, D. Sacca, D. Ursino.
An architecture for
accessing a large number of autonomous, heterogeneous databases.
{\em Networking and Information Systems Journal}, 1(4-5), 495-518, 1998.

\bibitem{levyprob}
D. Florescu, D. Koller, A. Levy.
Using Probabilistic Information for Data Integration.
{\em VLDB'97}.

\bibitem{flm}
D. Florescu, A. Levy, A. Mendelzon.
Database Techniques for the World-Wide Web: A survey.
{\em SIGMOD Record}, 1998.

\bibitem{flms}
D. Florescu, A. Levy, I. Manolescu, D, Suciu.
Qury Optimization in the Presence of Limited Access Paterns.
{\em SIGMOD'99}.

\bibitem{flm1}
M. Friedman, A. Levy, T. Millstein.
Navigational Plans for Data Integration.
{\em AAAI'99}.

\bibitem{razor}
M. Friedman, D. Weld. Efficient execution of information gathering plans.
{\em IJCAI'97}.

\bibitem{Freitag98}
D. Freitag.
Information Extraction from HTML: Application of a General
Machine Learning Approach. {\em AAAI'98}.

\bibitem{tsimmis1}
H. Garcia-Molina, Y. Papakonstantinou , D. Quass , A. Rajaraman , Y. Sagiv , J. Ullman , V.
Vassalos , J. Widom. The TSIMMIS approach to mediation: Data models and Languages.
{\em Journal of Intelligent Information Systems}, 8(1997).

\bibitem{gly}
H. Garcia-Molina, W. Labio, R. Yermeni. Capability sensitive query processing
on internet sources. {\em ICDE'99}.

\bibitem{VanGelder93}
A. van Gelder, K. Ross, J Schlipf.
The Well-Founded Semantics for General Logic Programs.
{\em Journal of the ACM}  38(3), 1991.

\bibitem{info1}
M. Genesereth, A Keller, O. Duschka. Infomaster: An Information Integration System.
{\em SIGMOD 97}.

\bibitem{glushko}
R. Glushko, J. Tenenbaum, B. Meltzer.
An XML framework for Agent-based E-commerce.
{\em Communications of the ACM}, Vol. 42, 1999.

\bibitem{rousset-dl00}
F. Goasdoue, M-C. Rousset.
Rewriting Conjunctive Queries using Views in Description Logics with
Existential Restrictions.
{\em Proceedings of the International Workshop on Description Logics, (DL'00)}, 2000.

\bibitem{picsel}
F. Goasdoue, V. Lattes, M-C. Rousset.
The use of CARIN language and algorithms for Information Integration: the
PICSEL project.
{\em International Journal on Cooperative Information Systems}, 9(4), 2000.


\bibitem{dataguide}
R. Goldman, J. Widom.
Dataguides: Enabling query formulation and optimization in
semistructured databases.
{\em VLDB'97}.

\bibitem{interact}
R. Goldman, J. Widom. Interactive Query and Search in Semistructured Databases.
{\em Proc. of the First
Intern. Workshop on the Web and Databases, WebDB'98}, LNCS 1590, 1998.

\bibitem{grosof99}
B. N. Grosof, Y. Labrou, H. Y. Chan.
A Declarative Approach to Business Rules in Contracts: Courteous Logic
Programs in XML.
{\em In Proceedings of the 1st ACM Conference on Electronic Commerce (EC-99)},
Denver, Colorado, USA, November, 1999.

\bibitem{Maes2}
R. Guttman, A. Moukas, and P. Maes.
Agent-mediated Electronic Commerce: A Survey.
{\em Knowledge Engineering Review}, 1998.

\bibitem{garlic1}
L. Haas, D. Kossmann, E. Wimmers, J. Yang.
Optimizing queries across diverse data sources.
{\em VLDB'97}

\bibitem{shoe}
J. Heflin, J. Hendler, S. Luke.
SHOE: A Knowledge Representation Language for
Internet Applications. {\em Technical Report CS-TR-4078} (UMIACS TR-99-71), Dept. of
Computer Science, University of Maryland at College Park. 1999.

\bibitem{OIL2}
F. van Harmelen, I. Horrocks.
FAQs on OIL: The Ontology Inference Layer.
{\em IEEE Intelligent Systems},  Vol. 15(6), p. 69-72, 2000.

\bibitem{DAML1}
J. Hendler, D. McGuinness.
The DARPA Agent Markup Language.
{\em IEEE Intelligent Systems},  Vol. 15(6), p. 72-73, 2000.

\bibitem{fact1}
I. Horrocks.
Using an expressive description logic: FaCT or fiction?
{\em KR'98}.

\bibitem{fact2}
I. Horrocks, U. Sattler, S. Tobies.
Practical reasoning for expressive description logics.
{\em Proceedings of the 6th International Conference on Logic for
 Programming and Automated Reasoning (LPAR'99)},
LNAI 1705, 1999.

\bibitem{hull}
R. Hull. Managing Semantic Heterogeneity in Databases: A Theoretical Pespective.
{\em PODS'97}.

\bibitem{ifflw}
Z. Ives, D. Florescu,M. Friedman, A. Levy, D. Weld.
An Adaptive Query Execution System for Data Integration.
{\em SIGMOD'99}.

\bibitem{webwatch}
T. Joachims, D. Freitag, T. Mitchell.
Webwatcher: A tour guide for the world wide web.
{`em IJCAI'97}, 1997.

\bibitem{KakasSurvey}
A.~C. Kakas, R.A. Kowalski, F.~Toni. The role of abduction in
logic programming. Handbook of Logic in Artificial Intelligence
and Logic Programming 5, pages 235-324, D.M. Gabbay, C.J. Hogger
and J.A. Robinson eds., Oxford University Press, 1998.

\bibitem{tonyvldb}
A. Kakas, P. Mancarella.
Database updates through abduction.
{\em Proc. 16th International Conference on Very Large Databases,
VLDB'90},1990.

\bibitem{sheth1}
V. Kashyap, A. Sheth.
Semantic Heterogeneity in Global Information Systems:
The Role of Metadata, Context and Ontologies. In
{\em Cooperative Information Systems: Current Trends and
         Directions}, Academic Press.

\bibitem{F-logic}
M. Kifer, G. Lausen, J. Wu. Logical Foundations of Object-Oriented
and Frame-Based Languages. {\em Journal of the ACM}, 42(4), 1995.

\bibitem{knoblock}
G. Knoblock.
Planning, Executing, Sensing and Replanning for Information Gathering.
{\em IJCAI'95}.

\bibitem{craigcia}
C. Knoblock, S. Minton, J. L. Ambite, N. Ashish, P. J. Modi, I. Muslea, A.
Philpot, S. Tejada.
Modeling Web Sources for Information Integration.
{\em AAAI'98}.


\bibitem{bargain}
B. Krulwich.
The BargainFinder Agent: Comparison price shopping on the Internet.
In {\em Agents, Bots, and other Internet Beasties}, 1996.

\bibitem{kushmerik}
N. Kushmerick, R. Doorenbos, and D. Weld. Wrapper Induction for
Information extraction. {\em IJCAI'97}.

\bibitem{SKC-project}
J. Jannink, P. Mitra, E. Neuhold, S. Pichai, R. Studer, G. Wiederhold.
An Algebra for Semantic Interoperation of Semistructured Data.
{\em IEEE Knowledge and   Data Engineering Exchange Workshop
(KDEX'99)}, 1999.

\bibitem{Junker}
M. Junker, M. Sintek, M. Rinck.
Learning for Text Categorization and Information Extraction
with ILP.
{\em Proc. Workshop on Learning Language in Logic}, ILLL99, Slovenia,
1999.

\bibitem{weblog}
LVS. Lakshmanan, F. Sadri, I.N. Subramanian. A Declarative
Approach to Querying and Restructuring the World-Wide-Web. {\em
Post - ICDE Workshop on Research Issues in Data Engineering
RIDE-96}, New Orleans, 1996.

\bibitem{lk}
E Lambrecht, S. Kambhampati.
Optimization Strategies for Information Gathering Plans.
{\em IJCAI'99}. Also available as ASU-CSE-TR 98-018.

\bibitem{RDF}
O. Lassila, R. Swick.
Resource Description Framework (RDF) Model and Syntax Specification.
http://www.w3.org/TR/1999/REC-rdf-syntax-19990222, 1999.

\bibitem{lreview}
A. Levy.
The Information Manifold approach to data integration.
In \cite{review}.

\bibitem{levy96}
A. Levy. Obtaining complete Answers from incomplete Databases.
{\em VLDB'96}.

\bibitem{levyrev99}
A. Levy.
Combining Artificial Intelligence and Databases for Data Integration
To appear in a {\em Special issue of LNAI on Artificial Intelligence Today;
Recent Trends and Developments}, 1999.

\bibitem{lro}
A. Levy, A. Rajaraman, J. Ordille.
Querying Heterogeneous Information Sources Using Source Descriptions.
{\em VLDB 96}.

\bibitem{levyullman99}
A. Levy, A. Rajaraman, J. Ullman.
Answering Queries Using Limited External Query Processors.
JCSS 58(1): 69-82, 1999.

\bibitem{levypatterns99}
A. Levy, I. Manolesu, D. Suciu, D. Florescu.
Query Optimization in the Presence of Limited Access Patterns.
{\em Proc. of ACM SIGMOD Conf. on Management of Data}, 1999.

\bibitem{lmss}
A. Levy, A. Mendelzon, Y. Sagiv, D. Srivastava.
Answering queries using views. {\em VLDB 96}.

\bibitem{lsk}
A. Levy, D. Srivastava, T. Kirk.
Data Model and Query Evaluation in Global Information Systems.
{\em Journal of Intelligent Information Systems}, 5, 1995.

\bibitem{levyreview00}
A. Levy. Answering Queries using Views: A Survey. Submitted, 2000.

\bibitem{Liu}
H. Li.
XML and Industrial Standards for Electronic Commerce.
{\em Knowledge and Information Systems}, Vol. 2, 2000.

\bibitem{letizia}
H. Lieberman.
Letizia: An Agent That Assists Web Browsing.
{\em IJCAI'95}, 1995.


\bibitem{multiDB1}
W. Litwin. From Database Systems to Multidatabase Systems: Why
and How. {\em BNCOD 1988}, pp. 161-188, 1988.

\bibitem{LogicWeb}
S.W. Loke, A. Davison. LogicWeb: Enhancing the Web withLogic
Programming, {\em Journal of Logic Programming} Vol.  36, pp.
195-240, 1998.

\bibitem{LogicWebTool}
S. W. Loke, A. Davison, L. Sterling. Lightweight DEDUCTIVE
databases on the World-Wide Web. {\em Proc. of the 1st Workshop on
Logic Programming Tools for Internet Application of Prolog},
London, April, 1996.

\bibitem{LogicWebTours}
S. W. Loke, A. Davison. A  Logic Programming Approach to
Generating Web-based Guided Tours. {\em Proc. of the 5the
Int.Conf. and Exhibitions on The Practical Application of Prolog},
London, April, 1997.


\bibitem{bertram}
B. Lud\"ascher, R. Himmerr\"oder, G. Lausen, W. May, C. Schlepphorst.
Managing Semistructured Data with Florid: A deductive Object-Oriented Perspective.
{\em Information Systems}, 23(1998).

\bibitem{shoe}
S. Luke, L. Spector, D. Rager, J. Hendler.
Ontology-based Web Agents. In {\em First International Conference on
Autonomous Agents}, 1997.

\bibitem{ieee-learn}
A. Maedche, S. Staab.
Ontology Learning for the Semantic Web.
{\em IEEE Intelligent Systems},  Vol. 16(2), p. 72-79, 2001.

\bibitem{Maes}
P. Maes, R. Guttman, A. Moukas.
Agents that Buy and Sell.
{\em Communications of the ACM}, Vol. 42, 1999.

\bibitem{metalog}
M. Marchiori, J. Saarela.
Query + metadata + logic = metadata.
{\em W3C Query Languages Workshop} ,1998.


\bibitem{MONDIAL}
W. May.
Information Retrieval and Integration with FLORID:
The Mondial Case Study. http://www.informatik.uni-freiburg.de/~may/Mondial/

\bibitem{WWWCM-99}
W. May, R. Himmer\"oder, G. Lausen,  B. Lud\"ascher.
A Unified Framework for Wrapping, Mediating and Restructuring Information
from the Web. {\em Proc. International Workshop on the World-Wide Web and
Conceptual Modeling, WWWCM'99}, 1999.

\bibitem{loom}
R, McGregor. The evolving technology of classification-based knowledge representation
systems. In {\em Principles of Semantic Networks: Explorations in the Representation
of Knowledge}, J. Sowa (ed.), 1990.

\bibitem{deborah}
D. L. McGuinness.
Ontologies and Online Commerce.
{\em IEEE Intelligent Systems},  Vol. 16(1), pp. 9-10, 2001.

\bibitem{lorel2}
J. McHugh, S. Abiteboul, R. Goldman, D. Quass, J. Widom.
Lore: A Database Management System for Semistructured Data.
{\em SIGMOD Record}, 26, 1997.

\bibitem{DAML2}
S. McIlraith, T.C. Son, H. Zeng.
Semantic Web Services.
{\em IEEE Intelligent Systems},  Vol. 16(2), p. 46-53, 2001.

\bibitem{observ99}
E. Mena, A. Illarramendi, V. Kashyap, A. Sheth.
OBSERVER: An Approach for Query Processing in
Global Information Systems based on Interoperation
across Pre-existing Ontologies. {\em Distributed and Parallel Databases Journal},
8(2), p. 223-271, 2000.

\bibitem{milo}
T. Milo, S. Zohar.
Using schema matching to simplify heterogeneous data translation.
{\em VLDB'98}.

\bibitem{LB-AI-99}
J. Minker (editor). Logic-Based Artificial Intelligence. Kluwer
Academic Publishers, 2000.

\bibitem{minsky}
M. Minsky.
Commonsense-Based Interfaces.
{\em Communications of the ACM}, 43(8), 67-73, 2000.

\bibitem{ieee-mine}
B. Mobasher, R. Cooley, J. Srivastava.
Automatic Personalization Based on Web Usage Mining.
{\em Communications of the ACM}, 43(8), 142-151, 2000.

\bibitem{D3web}
A. Montes, J.F and Yague del Valle, M.I.  "Querying the Web with
Higher Expressive Power: D3Web", Dpt. Lenguajes y Ciencias de la
Computacion. Univ. de Malaga. Spain, 1997.


\bibitem{Maes1}
A. Moukas, R. Guttman, P. Maes.
Agent-mediated Electronic Commerce: An
MIT Media Laboratory Perspective.
{\em International Conference on Electronic Commerce}, 1998.

\bibitem{muslea-intro}
I. Muslea.
Extraction Patterns for Information Extraction Tasks: A Survey.
{\em AAAI-99 Workshop on Machine Learning for Information Extraction}, 1999.

\bibitem{carnot}
Munindar P. Singh, Philip Cannata, et al.
The Carnot Heterogeneous Database Project: Implemented Applications.
Distributed and Parallel Databases 5(2): 207-225 (1997) 1993.

\bibitem{NAM}
S. Nestorov, S. Abiteboul, R. Motwani.
Inferring structure in semistructured data.
{\em Workshop on Management of Semistructured Data}, 1997.

\bibitem{nestorov1}
S. Nestorov, J. Ullman, J. Weiner, S. Chawathe.
Representative objects: Concise representation of semistructured hierarchical data.
{\em ICDE'97}.

\bibitem{nestorov}
S. Nestorov, S. Abiteboul, R. Motwani.
Extracting Schema from Semistructured Data.
{\em SIGMOD'98}.

\bibitem{Ng}
W.K. Ng, E.P. Lim.
Standarization and Integration in Business-to-Business
Electronic Commerce.
{\em IEEE Intelligent Systems},  Vol. 16(1), pp. 12-14, 2001.

\bibitem{sleuth2}
M. Nodine, W. Bohrer, A. H. Hiong Ngu.
Semantic Brokering over Dynamic Heterogeneous Data Sources in InfoSleuth.
{\em ICDE'99}.

\bibitem{borys}
B. Omelayenko.
Learning of Ontologies for the Web: the Analysis of Existent Approaches.
{\em Proc. International Workshop on Web Dynamics}, London, 2001.

\bibitem{ddbs-carnot95}
K. Ong, N. Arni, et al.
A Deductive Database Solution to Intelligent Information Retrieval
from Legacy Databases.
DASFAA 172-179, 1995.

\bibitem{ouksel}
A, Ouksel, A. Sheth. Semantic Interoperability in Global Information Systems.
{\em SIGMOD Record} 28(1), 1999.

\bibitem{calabria-scheme-1}
L. Palopoli, D. Sacca, G. Terracina, D. Ursino.
A unified graph-based framework for deriving nominal
interscheme properties, type conflicts and object cluster similarities.
{\em 4th IFCIS Conference on Cooperative
Information Systems (CoopIS'99)}, Edimburgh (UK), IEEE Press, 34-45, 1999.

\bibitem{calabria-scheme-2}
 L. Palopoli, D. Sacca, D. Ursino.
 Semi-automatic techniques for deriving
interscheme properties from database schemes.
{\em Data and Knowledge Engineering}, 30(3), 239-273, 1999.

\bibitem{calabria-scheme-3}
 L. Palopoli, D. Sacca, D. Ursino.
 $DL_P$: A description logic for
extracting and managing complex terminological and structural properties
from database schemes.
{\em Information Systems}, 24(5), 403-425, 1999.

\bibitem{ineel}
B. Panchapagesan, J. Hui, G. Wiederhold, S. Erickson, L. Dean.
The INEEL Data Integration Mediation System.
http://id.inel.gov/idim/paper.html

\bibitem{pgh}
Y. Papakostantinou, A. Gupta, L. Haas. Capabilities-based query answering in
mediator systems. {\em Conference on Parallel and Distributed Information Systems,
PDIS}, 1996.

\bibitem{yannis-semi-re}
Y. Papakostantinou, V. Vassalos.
Query Rewriting for Semistructured Data.
{SIGMOD'99}.

\bibitem{parsons98}
R. Parson. Intelligent agents for the world wide web.
{\em Msc Thesis, Oxford University Computing Laboratory}, 1995.

\bibitem{PE95}
M. Perkiwitz, O. Etzioni. Category Translation: Learning to understand information
on the internet. {\em IJCAI'95}.

\bibitem{pohl}
W. Pohl, I. Schwab, I. Koychev.
Learning About the User: A General Approach and Its Application.
{\em IJCAI'99 Workshop ``Learning About Users''}, 1999.

\bibitem{rsu}
A. Rajaraman, Y. Sagiv, J. Ullman.
Aswering queries using templates with binding patterns.
{\em PODS pp. 105-112}, 1995.

\bibitem{XQL}
J. Robie, J. Lap, D. Schach.
XML Query Language (XQL).
{\em W3C Workshop on Query Languages}, 1998.

\bibitem{datalogforXML}
J. Shanmugasundaram, K. Tufte, G. He, C. Zhang, D. DeWitt, J.
Naughton. Relational Databases for Querying XML Documents:
Limitations and Opportunities, {\em Proc. of the 25th VLDB
Conference}, Edinburgh, Scotland, 1999.

\bibitem{firefly}
U. Shardanand and P. Maes.
Socail Information Filtering: Algorithms for Automating
``Word of Mouth''.
{\em Human Computer Interaction Conference, HCI'95}, 1995.

\bibitem{fedDB1}
A. P. Sheth, J. A. Larson. Federated Database Systems for Managing
Distributed, Heterogeneous, and Autonomous Databases. {\em ACM
Computing Surveys}, Vol. 22(3), pp. 183-236, 1990.

\bibitem{dbs-use95}
A. P. Sheth, C. Wood, V. Kashyap.
Q-Data: Using Deductive Database Technology to Improve Data Quality.
Applications of Logic Databases (R. Ramakrishnam ed.), Kluwer Academic, 1995.

\bibitem{smith}
H. Smith, K. Poulter.
Share the Ontology in XML-based Trading Architecture.
{\em Communications of the ACM}, Vol. 42, 1999.

\bibitem{Soderland97}
S. Soderland. Learning to extract text-based information from the world wide web.
{\em Proceedings of the 3rd International Conference on Knowledge Discovery and Data
Mining}, 1997.

\bibitem{srivastava}
J. Srivastava, R. Cooley, M. Deshpande, P. Tan.
Web Usage Mining: Discovery and Applications of Usage Patterns form
Web Data.
{\em SIGKDD Explorations}, 1(2), ACM, 2000.

\bibitem{ecaiws}
S. Staab, A. Maedche, C. Nedellec, P. Wiemer-Hastings.
{\em Proc. of the Ontology Learning ECAI-2000 Workshop},
http://ol2000.aifb.uni-karlsruhe.de, 2000.

\bibitem{ijcaiws}
S. Staab, A. Maedche, C. Nedellec, E. Hovy.
{\em Proc. of the Ontology Learning IJCAI-2001 Workshop},
http://ol2001.aifb.uni-karlsruhe.de, 2001.

\bibitem{suciureview}
D. Suciu.
An overview of Semistructured Data.
{\em SIGACT News}, 29(4), 1998.

\bibitem{dataware2}
D. Theodoratos, T. K. Sellis. Designing Data Warehouses. {\em DKE}
Vol. 31(3), pp. 279-301, 1999.

\bibitem{fedDB2}
G. Thomas, G. R. Thompson, Chin-Wan Chung, E. Barkmeyer, F.
Carter, M. Templeton, S. Fox, B. Hartman. Heterogeneous
Distributed Database Systems for Production Use. {\em ACM
Computing Surveys}, Vol. 22(3), pp. 237-266, 1990.

\bibitem{disco}
A, Tomasic, L. Raschid, P. Valduriez. Scaling
Heterogeneous Databases and the Design of Disco.
{\em INRIA Report}, No 2704, 1995.

\bibitem{vp}
V. Vassalos, Y. Papakostantinou. Describing and using query capabilities of
heterogeneous sources. {\em VLDB 97}.

\bibitem{vp1}
V. Vassalos, Y. Papakostantinou.
Expressive Capabilities Description Languages and Query Rewriting Algorithms.
{\em Journal of Logic Programming}, to appear.

\bibitem{vp2}
V. Vassalos, Y. Papakostantinou.
 Using Knowledge of Redundancy for Query Optimization in Mediators.
{\em Workshop on AI and Information Integration}, AAAI'98, 1998.

\bibitem{ullman97}
J. Ullman. Information Integration Using Logical Views. {\em ICDT'97}.

\bibitem{kraft}
P.R.S. Visser, D.M. Jones, M. Beer, T.J.M. Bench-Capon, B. Diaz, M.J.R. Shave.
Resolving Ontological Heterogeneity in the KRAFT Project.
{\em DEXA-99}, Florence, Italy, August 30-September 3, 1999.

\bibitem{dataware3}
J. Widom.
 Research Problems in Data Warehousing, {\em Proc. of ACM
CIKM'95}, pp. 25-30, 1995.

\bibitem{wie92}
G. Wiederhold. Mediators in the Architecture of Future Information System.
{\em IEEE Computer}, 25(3), 1992.

\bibitem{wie94}
G. Wiederhold. Interoperation, Mediation and Ontologies.
{\em Fifth Generation Computer Systems, FGCS'94}.

\bibitem{wie97}
G. Wiederhold, M. Genesereth.
The Conceptual Basis for Mediation Services. {\em IEEE Expert}, 12, 1997.

\bibitem{giofuture}
G. Wiederhold, R. Jiang.
Information Systems That Really Support Decision-Making.
{\em ISMIS 99}, Springer LNCS, 1999.

\bibitem{widomdirect}
J. Widom.
Data Management for XML: Research Directions.
{\em IEEE Data Engineering Bulletin},22(3), 1999

\bibitem{xanthakos}
Ionnis Xanthakos. Semantic integration of information and logic-based agents.
{\em MSc Advanced Computing thesis}, Department of Computing, Imperial
College, 1999.

\bibitem{XML}
http://www.w3.org/XML/

\bibitem{yl}
 R. Yerneni, C. Li.
 Optimizing large join queries in mediaton systems.
 {\em ICDT'99}.

\bibitem{ylgu}
R. Yerneni, C. Li, H. Garcia-Molina, J. Ullman.
Computing Capabilities of Mediators.
{\em SIGMOD' 99}.

\bibitem{ldl++}
C. Zaniolo.
A Short Overview of LDL++: A Second-Generation Deductive Database System.
{\em Computational Logic}, Vol 3, No. 1, pp. 87-93, Dec. 1996.

\bibitem{housing}
http://logic.stanford.edu/rentals/

\bibitem{federal}
http://infomaster.stanford.edu:4500/

\bibitem{flight}
http://www.isi.edu/ariadne/demo/tw/index.html

\bibitem{datajoiner}
http://www.software.ibm.com/data/datajoiner/brochure/index.html

\bibitem{omni}
http://www.sybase.com/products/entcon/omni.html.

\bibitem{crossaccess}
http://www.crossaccess.com.

\bibitem{enterworks}
http://www.enterworks.com/products/products\_vdb.html.

\bibitem{OIL}
http://www.ontoknowledge.org/oil/index.shtml.

\bibitem{DAML}
http://www.daml.org.

\bibitem{DAML+OIL}
http://www.daml.org/2001/03/daml+oil-index.

\bibitem{TZI-DE}
http://www.tzi.de/grp/i3/.

\bibitem{DUSKA-Projects}
http://logic.stanford.edu/people/duschka/projects.html.

\bibitem{UNI-KARL-DE}
http://www.aifb.uni-karlsruhe.de/WBS/broker/inhalt-wp.html.

\bibitem{Interprice}
www.interprice.com.

\bibitem{ContentEurope}
www.contenteurope.com.

\bibitem{GrosofIBM}
http://alphaworks.ibm.com.

\end{thebibliography}
\end{document}